%% file: Formatting-Instructions-LaTeX-2026.tex
\documentclass[letterpaper]{article} % DO NOT CHANGE THIS
\usepackage{aaai2026}  % DO NOT CHANGE THIS
\usepackage{times}  % DO NOT CHANGE THIS
\usepackage{helvet}  % DO NOT CHANGE THIS
\usepackage{courier}  % DO NOT CHANGE THIS
\usepackage[hyphens]{url}  % DO NOT CHANGE THIS
\usepackage{graphicx} % DO NOT CHANGE THIS
\usepackage{natbib}  % DO NOT CHANGE THIS AND DO NOT ADD ANY OPTIONS TO IT
\usepackage{caption} % DO NOT CHANGE THIS AND DO NOT ADD ANY OPTIONS TO IT
\usepackage{algorithm}
\usepackage{algorithmic}

\usepackage{amssymb}
\usepackage{bbding}

\usepackage{booktabs}
\usepackage{multirow}
\usepackage{adjustbox}
\usepackage{caption}

\usepackage{makecell}
\usepackage{amsmath}
\usepackage{pifont}
\usepackage[most]{tcolorbox}
\usepackage[table]{xcolor}

\usepackage{newfloat}
\usepackage{listings}

\def\eg{\emph{e.g.}}
\def\ie{\emph{i.e.}}

\usepackage{newfloat}
\usepackage{listings}
\DeclareCaptionStyle{ruled}{labelfont=normalfont,labelsep=colon,strut=off} % DO NOT CHANGE THIS
\floatstyle{ruled}
\newfloat{listing}{tb}{lst}{}
\floatname{listing}{Listing}
\definecolor{myred}{RGB}{162, 29, 27}
\newcounter{mycounter} % create a new counter, called 'mycounter'

\newcommand{\findingbox}[1]{
    \stepcounter{mycounter} % Increment counter
    \begin{tcolorbox}[colframe=black,
                      arc=1pt,
                      boxsep=-2pt,
                      ]
        \noindent{\textcolor{myred}{\textbf{\textit{Finding \themycounter.}}}} #1
    \end{tcolorbox}
}

\title{LLMC+: Benchmarking Vision-Language Model Compression with a Plug-and-play Toolkit}
\author{
    Chengtao Lv\textsuperscript{\rm 1, 2}\thanks{Work done during internships at SenseTime Research.}, Bilang Zhang\textsuperscript{\rm 2, 3$*$}, Yang Yong\textsuperscript{\rm 2}, Ruihao Gong\textsuperscript{\rm 2, 3\textdagger}, Yushi Huang\textsuperscript{\rm 2, 4$*$}, \\
    Shiqiao Gu\textsuperscript{\rm 2}, Jiajun Wu\textsuperscript{\rm 3}, Yumeng Shi\textsuperscript{\rm 1}, Jinyang Guo\textsuperscript{\rm 3}, Wenya Wang\textsuperscript{\rm 1}\thanks{Corresponding authors.}
}
\affiliations{
    \textsuperscript{\rm 1}Nanyang Technological University, \textsuperscript{\rm 2}SenseTime Research\\
    \textsuperscript{\rm 3}Beihang University, \textsuperscript{\rm 4}Hong Kong University of Science and Technology\\
    \small{\texttt{\{chengtao001, yumeng001\}@e.ntu.edu.sg, wangwy@ntu.edu.sg}}\\
    \small{\texttt{\{zhangbilang, yongyang, gongruihao, huangyushi1, gushiqiao\}@sensetime.com}}
}

\usepackage{bibentry}
\begin{document}

\maketitle

\begin{abstract}
Large Vision-Language Models (VLMs) exhibit impressive multi-modal capabilities but suffer from prohibitive computational and memory demands, due to their long visual token sequences and massive parameter sizes. To address these issues, recent works have proposed training-free compression methods. However, existing efforts often suffer from three major limitations: (1) Current approaches do not decompose techniques into comparable modules, hindering fair evaluation across spatial and temporal redundancy. (2) Evaluation confined to simple single-turn tasks, failing to reflect performance in realistic scenarios. (3) Isolated use of individual compression techniques, without exploring their joint potential. To overcome these gaps, we introduce LLMC+, a comprehensive VLM compression benchmark with a versatile, plug-and-play toolkit. LLMC+ supports over 20 algorithms across five representative VLM families and enables systematic study of token-level and model-level compression. Our benchmark reveals that: (1) Spatial and temporal redundancies demand distinct technical strategies. (2) Token reduction methods degrade significantly in multi-turn dialogue and detail-sensitive tasks. (3) Combining token and model compression achieves extreme compression with minimal performance loss. We believe LLMC+ will facilitate fair evaluation and inspire future research in efficient VLM. Our code is available at \url{https://github.com/ModelTC/LightCompress}.
\end{abstract}


\begin{figure*}[!t]
\begin{center}
     \includegraphics[width=0.98\linewidth]{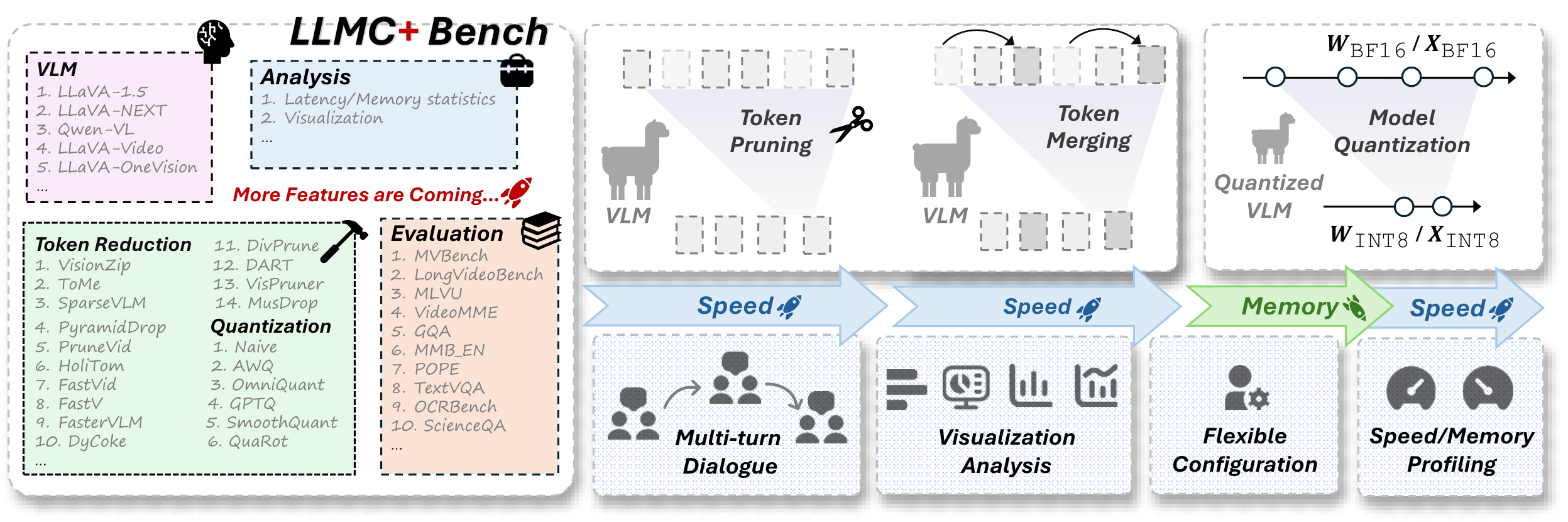}
\end{center}
  \caption{Illustration of our proposed powerful toolkit, LLMC+. Due to its high flexibility and versatility, we build a VLM compression benchmark upon it and conduct an in-depth analysis.}
    \label{fig:frame}
\end{figure*}

\section{Introduction}

Recently, Large Language Models (LMMs)~\cite{touvron2023llama,liu2024deepseek,brown2020language} have achieved rapid advancements in Natural Language Processing (NLP), which has become a significant milestone in the AI revolution. This breakthrough has quickly extended to vision modalities: mainstream Vision Language Models (VLMs)~\cite{liu2023visual,liu2024improved,wang2024qwen2,chen2024internvl} typically encode visual inputs into tokens and unify multiple modalities within a shared embedding space, demonstrating strong visual-language understanding and generation capabilities in various tasks~\cite{singh2019towards,antol2015vqa,hudson2019gqa}. 

% Recently, Vision Language Models (VLMs)~\cite{liu2023visual,liu2024improved,wang2024qwen2,chen2024internvl} have achieved rapid advancements in Natural Language Processing (NLP). Mainstream Vision Language Models typically encode visual inputs into tokens and unify multiple modalities within a shared embedding space, demonstrating strong visual-language understanding and generation capabilities in various tasks~\cite{singh2019towards,antol2015vqa,hudson2019gqa}. 

However, their remarkable capabilities can be largely attributed to two aspects: 1) The number of visual tokens often reaches hundreds or even thousands, dominating the input. For example, images in LLaVA-NeXT~\cite{liu2024llavanext} are converted into 2,880 tokens. While in video streams or high-resolution scenarios, the number of tokens increases dramatically, further intensifying the computational costs. 2) VLMs have massive memory footprints (\eg, billion-scale parameters). Some large-scale VLMs, such as Qwen2.5-VL-72B~\cite{wang2024qwen2}, consume approximately 140GB of memory for storage, becoming a major GPU memory bottleneck during inference. These two issues constrain their widespread application on resource-limited devices.

To effectively mitigate intractable computational and memory overhead, several training-free compression works have been proposed successively, which can be briefly classified into two fields: 1) \textit{token-level compression}. These methods typically reduce less salient visual tokens through token reduction~\cite{bolya2022token,chen2024image}. 2) \textit{model-level compression}. Their primary objective is to squeeze model weights through techniques like quantization~\cite{gong2025survey},  network pruning~\cite{sun2023simple}, and low-rank factorization~\cite{wang2024svd}.

% First, existing methods often focus on different technical dimensions, but there is a lack of fair comparison and in-depth investigation across these dimensions. 
Nevertheless, three worrisome problems still appear in current training-free compression research for VLMs. First, existing methods often target different types of redundancy (\eg, spatial or temporal) with distinct technical dimensions, leading to a lack of fair comparison and in-depth analysis across these dimensions. Second, these methods are limited to evaluation on general single-turn VQA tasks, lacking comprehensive assessments on challenging and practical tasks. Third, they typically rely on a single compression measure, without exploring the risks and potential of joint multiple compression strategies.
% Based on LLMC+, we distill three core technical dimensions: applying different metrics (attention-based vs. similarity-based) and techniques (merge vs. prune) across distinct model stages (Vision Tower vs. LLM).
% We then conduct extensive experiments and perform solid analysis along these dimensions.

 To this end, this paper presents LLMC+, a VLM compression benchmark with a versatile toolkit covering \textit{token-level} and \textit{model-level} compression. Specifically, LLMC+ supports over 20 compression algorithms and five different families of VLMs. 1) Based on LLMC+, we introduce a novel token reduction taxonomy specifically for handling spatial and temporal redundancy. We further distill their core technical dimensions, covering metrics (\textit{attention-based vs. similarity-based}), solutions (\textit{merge vs. prune}), and video segmentation strategies (\textit{fixed vs. dynamic}). Extensive experiments are conducted along these dimensions, accompanied by in-depth analysis. 2) Besides, we evaluate VLMs on practical task scenarios, such as multi-turn dialogue and detail-sensitive tasks like OCR and DocVQA, to uncover potential risks introduced by compression. 3) Finally, we combine token reduction and quantization to achieve extreme compression. We deploy quantized kernels on hardware to validate real acceleration and memory saving. Our benchmark reveals that 1) Spatial and temporal redundancy require distinct core strategies to be handled effectively. 2) Token reduction methods suffer from non-trivial performance degradation on practical tasks. 3) Combining multiple techniques enables extreme compression with accuracy guarantees.

We emphasize that the contributions of our LLMC+ can be summarized as follows:
\begin{itemize}

% 1. Versatile toolkit: xxx
% 2. Modular comparison: xxx
% 3. Pratical evaluation: xxx
% 4. Best practice: xxx

% \item \textit{LLMC+ is the \textbf{first} versatile, plug-and-play compression toolkit tailored for VLMs.} LLMC+ supports dozens of algorithms and models, enabling developers and researchers to analyze VLM compression with flexible configuration.
% \item \textit{Our systematic benchmark provides a brand new perspective of VLM compression.} We conduct quantitative evaluations across a wide range of models and tasks based on diverse core technical dimensions.

\item \textit{Versatile Toolkit.} LLMC+ is the \textbf{first} plug-and-play compression toolkit specifically designed for VLMs, supporting over 20 compression algorithms across five different families of VLMs with flexible configuration.

\item \textit{Modular Comparison.} We construct a comprehensive taxonomy for token reduction that comprises all core technical modules, and conduct systematic evaluations for each module to ensure fair comparison.

\item \textit{Practical Evaluation.} Our findings highlight flaws in current evaluation practices and advocate for integrating our proposed practical evaluations into future evaluation standards.

\item \textit{Best Practice.} By following the modular guidelines proposed in this paper and combining token reduction with quantization, we achieve extreme compression.

\end{itemize}

\section{LLMC+ Implementation}
% To begin with, we develop a versatile compression toolkit for VLMs. We highlight several key features in Fig.~\ref{fig:frame}.
To enable comprehensive and fair comparison, we develop a versatile compression toolkit for VLMs. We highlight several key features in Fig.~\ref{fig:frame}.

\subsection{Various Algorithms and Models} 
LLMC+ supports a wide range of compression schemes, which can be broadly categorized into two fields. The first category includes 15 token reduction algorithms that concentrate on accelerating inference speed. The second category comprises model compression methods, covering 6 quantization algorithms, including both weight-only and weight-activation quantization. Moreover, LLMC+ integrates VLMs from different families, ranging from traditional image-based VLMs to video-oriented ones, including LLaVA-1.5~\cite{liu2023visual}, LLaVA-NeXT~\cite{liu2024llavanext}, Qwen2.5-VL~\cite{bai2025qwen2}, Video-LLaVA~\cite{lin2023video}, and LLaVA-OneVision~\cite{li2024llava}.

\subsection{Flexible Configuration}
The modular design of LLMC+ facilitates modality-aware compression (\eg, Vision Tower and LLM), seamless integration of diverse compression techniques (\eg, token reduction and quantization), multi-turn dialogues, and convenient configuration for latency/memory profiling and visualization. LLMC+ enables developers to perform customized analysis and compression tailored to their specific needs.

\subsection{Benchmarks} LLMC+ is integrated with LMMs-Eval~\cite{zhang2024lmms} for evaluation. For image tasks, we conduct experiments on nine widely used image understanding benchmarks, including visual question answering benchmarks such as GQA~\cite{hudson2019gqa}, ScienceQA~\cite{saikh2022scienceqa}, TextVQA~\cite{singh2019towards}, and VizWiz~\cite{bigham2010vizwiz}, as well as multi-modal reasoning benchmarks such as MMBench~\cite{liu2024mmbench}, MME~\cite{fu2023mme}, POPE~\cite{li2023evaluating}, OCRBench~\cite{liu2024ocrbench}, and DocVQA~\cite{tito2023hierarchical}. For video tasks, we select four standard video understanding benchmarks: MVBench~\cite{li2024mvbench}, LongVideoBench~\cite{wu2024longvideobench}, MLVU~\cite{zhou2024mlvu}, and VideoMME~\cite{fu2025video}. These benchmarks cover complex scenarios and varying durations, enabling a comprehensive evaluation of effectiveness and generalization ability for these methods.

% \section{LLMC+ Evaluation and Analysis}
% \section{Explore Across Three Technical Dimensions}
\input{table/classification}

\section{Dive into Token Reduction}

% In this section, we first introduce our taxonomy of token reduction, which covers three core technical aspects. We then conduct experiments on the vision and language components separately. Finally, we focus on video-based tasks and explore key techniques specific to video.
In this section, we first present our taxonomy for addressing spatial and temporal redundancy (Sec.~\ref{sec:tax}), which comprises two core technical aspects, respectively. For spatial redundancy, we conduct experiments in the vision (Sec.~\ref{sec:red_v}) and language components (Sec.~\ref{sec:red_llm}) separately. For temporal redundancy, we distill a two-step pipeline and evaluate key techniques at each step (Sec.~\ref{sec:i_v}).

\subsection{Taxonomy}
\label{sec:tax}

% Tab.~\ref{tab:classify} presents the taxonomy of token reduction methods. We emphasize that this taxonomy encompasses all the token reduction approaches discussed in this paper. Based on different perspectives, we divide them into the following categories:
Token reduction primarily aims to eliminate redundancy in visual inputs. While images exhibit \textit{spatial redundancy}, videos additionally contain \textit{temporal redundancy}. 

First, we present a taxonomy of methods designed to address spatial redundancy, as demonstrated in Tab.~\ref{tab:classify}. Based on different perspectives, we divide them into the following categories:

% (Tab.~\ref{tab:classify_video})
\begin{itemize}

\item \textit{Attention-based vs. Similarity-based metric}. Attention-based metrics typically rely on the question prompt or the [\texttt{CLS}] token, while similarity-based metrics measure the pairwise distance between tokens.

\item \textit{Prune vs. Merge.} Pruning discards insignificant visual tokens, whereas merging fuses them into other tokens.

% \item \textit{Vision vs. Language.} Vision-level token reduction is performed within Vision Tower, while language-level token reduction occurs within the layers of the LLM.
\end{itemize}

Similarly, we categorize methods (see Tab.~\ref{tab:classify_video}) addressing temporal redundancy into the following types:

\begin{itemize}

\item \textit{Fixed vs. Dynamic segmentation}. Fixed segmentation partitions frames into segments of equal length, whereas dynamic segmentation generates segments that may vary in length.

\item \textit{Prune vs. Merge.} Similar to the above, but primarily applied to consecutive frames.

\end{itemize}

\input{table/vision_tower}

\subsection{Token Reduction in Vision Tower}\label{sec:red_v}

We report the detailed results of token reduction schemes for Vision Tower in Tab.~\ref{tab:vision_tower_main}. It is worth noting that for each row, like ``VisionZip PA" indicates the use of an \textbf{A}ttention-based metric to \textbf{P}rune the unimportant tokens. We have several key observations: (1) \textit{Prune-based methods generally outperform Merge-based ones in Vision Tower.} For example, three prune-based methods significantly outperform ``ToMe MS" by a large margin. (2) \textit{Similarity-based and Attention-based metrics, such as VisionZip PA~\cite{yang2025visionzip} and DivPrune PS~\cite{alvar2025divprune}, exhibit only trivial differences in accuracy for most settings.} In detail, [\texttt{CLS}] attention serves as a strong indicator of token importance within the encoder and has demonstrated superior performance compared to other methods, as evidenced by its adoption in several previous works~\cite{yang2025visionzip,zhang2024beyond,liu2024multi}. However, for some advanced Vision Towers that cannot obtain [\texttt{CLS}] attention, such as SigLIP~\cite{zhai2023sigmoid}. We believe there are two possible solutions. The first is to apply a similarity-based token pruning method, such as DivPrune PS~\cite{alvar2025divprune}. The second is to reintegrate the SigLIP head into the model to generate [\texttt{CLS}] token, which incurs negligible computational overhead.

% We observe that prune-based methods generally outperform merge-based ones.

Besides, some methods apply token reduction in the shallow layers of Vision Tower. For instance, MustDrop MS~\cite{liu2024multi} performs spatial merge with sliding windows in the first layer of Vision Tower, while ToMe MS~\cite{bolya2022token} adopts a progressive token reduction strategy, where tokens are gradually pruned layer by layer during forward inference. To compare the impact of applying these methods at different depths, we measure their performance when applied to either shallow or deep layers of Vision Tower. Tab.~\ref{tab:vision_tower_pos} shows that applying token reduction to shallow layers leads to significant accuracy degradation, while offering slight improvements in prefill time. Therefore, \textit{we recommend performing token reduction at the last layer of Vision Tower for a better trade-off between efficiency and accuracy.} However, these methods are not applicable to models like Qwen2.5-VL~\cite{bai2025qwen2}, because they occur before the model’s built-in spatial token merger module.

% \textit{Similar to the observations in Vision Tower, prune-based methods (i.e., FastV PA~\cite{chen2024image}, SparseVLM PA~\cite{zhang2024sparsevlm}, and DART PS~\cite{wen2025stop}) also surpass merge-based schemes (i.e., HoliTom MS~\cite{shao2025holitom}) in the LLM (Tab.~\ref{tab:llm_main}).} 
\input{table/llm}

\subsection{Token Reduction in LLM}\label{sec:red_llm}
In contrast to text-agnostic token reduction methods applied within Vision Tower, token reduction within the LLM often leverages the question prompt as guidance for identifying important tokens. These methods typically perform token reduction in the shallow layers of the LLM. For a fair comparison, we implement all methods in our taxonomy (Tab.~\ref{tab:classify}) at 5-th layer of the LLM during assessment. \textit{The effectiveness of methods in LLM varies substantially across model families.} For instance, within the LLaVA family, prune-based approaches (i.e., FastV PA~\cite{chen2024image}, SparseVLM PA~\cite{zhang2024sparsevlm}, and DART PS~\cite{wen2025stop}) consistently outperform merge-based schemes (e.g., HoliTom MS~\cite{shao2025holitom}). In contrast, for Qwen2.5-VL, HoliTom MS~\cite{shao2025holitom} demonstrates competitive performance. Token reduction methods applied within the LLM typically introduce additional computational overhead but tend to achieve better performance compared to those applied in Vision Tower. For instance, DART PS~\cite{wen2025stop} can maintain nearly lossless performance on LLaVA-1.5-7B even when preserving only one-third of the visual tokens. \textit{This also highlights a key limitation of attention-based metrics within the LLM.} They tend to assign higher attention scores to visual tokens that are spatially closer to text tokens~\cite{zhang2024beyond} (See in the Appendix). Therefore, it is necessary to reconsider the use of attention-based metrics for token reduction in LLM.

% Besides, FastV~\cite{chen2024image} also yields subpar accuracy, as it leverages last-token attention, which tends to preserve visual tokens that are spatially closer to the final token~\cite{zhang2024beyond}. 
% \input{table/prune_merge}
% \subsection{Joint Prune and Merge}

After comparing prune and merge algorithms in both the vision and language components, we further investigate whether combining token prune and merge can lead to improved performance. Recent works~\cite{yang2025visionzip,zhang2024sparsevlm} usually follow a two-step pipeline to integrate prune and merge: important tokens are first preserved through pruning, and then the remaining less informative contextual tokens are merged. Therefore, we select a representative algorithm from each of the vision and language parts for our experiments. As shown in Tab.~\ref{tab:vision_tower_main} and Tab.~\ref{tab:llm_main}, the standalone prune method and the combination (\ie, PA+MS) are compared. It can be observed that in most cases, introducing merge in addition to pruning does not significantly improve the model’s performance, and in some settings, it even leads to slightly lower accuracy compared to prune alone.

\begin{figure}[t]
\begin{center}
     \includegraphics[width=1\linewidth]{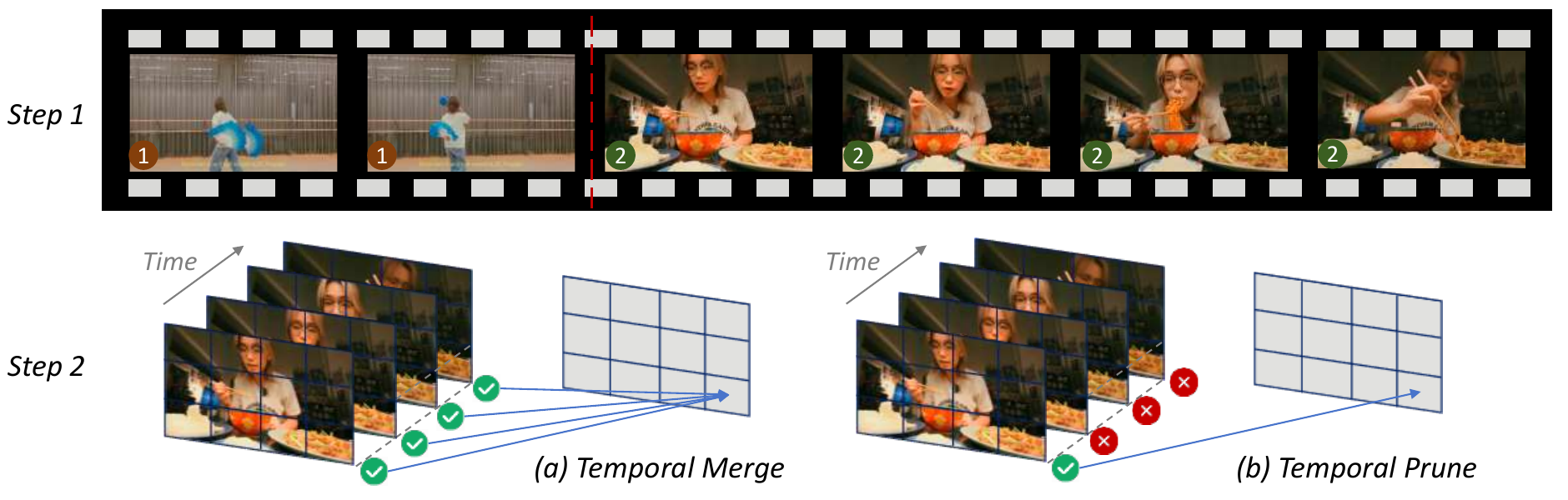}
\end{center}
  \caption{The pipeline of removing temporary redundancy in the two steps.}
    \label{fig:video_steps}
\end{figure}

\input{table/segment}

\subsection{From Image to Video}\label{sec:i_v}
Although several token reduction methods for spatial redundancy have been discussed in the aforementioned sections, directly applying these solutions to video tasks overlooks a key characteristic of video data that necessitates a rethinking of traditional pruning strategies. Compared with images, videos naturally introduce an additional form of considerable redundancy: temporal redundancy. This redundancy primarily arises from similar tokens in adjacent frames, which fail to provide additional informative content. Therefore, recent studies~\cite{tao2025dycoke,fu2024framefusion,shao2025holitom,huang2024prunevid,shen2025fastvid} aim to reduce such redundancy.

% Specifically, these approaches tailored for videos mainly follow a two-step paradigm, as shown in Fig.~\ref{fig:video_steps}. In the first step, these algorithms partition a video into temporally ordered segments with high similarity, which often correspond to the same scene. In the second step, they reduce unimportant visual tokens within each segment while preserving the informative ones. The remainder of this section explores how to perform these two steps to effectively reduce temporal redundancy.
Specifically, these approaches tailored for videos mainly follow a two-step paradigm (Fig.~\ref{fig:video_steps}). In the first step, these algorithms partition a video into temporally ordered segments with high similarity, which often correspond to the same scene. In the second step, they reduce unimportant visual tokens within each segment while preserving informative ones. The remainder of this section explores how to perform these two steps to effectively reduce temporal redundancy.

\textbf{First step:} To thoroughly explore different segment partition strategies, we summarize four representative methods, covering DyCoke~\cite{tao2025dycoke}, FastVID~\cite{shen2025fastvid}, PruneVid~\cite{huang2024prunevid}, and HoliTom~\cite{shao2025holitom}. To ensure a fair comparison, we apply a uniform merge strategy to highly similar tokens within each segment after partitioning. In addition to reporting performance under comparable compression ratios, we also provide the segmentation time and overall prefill time.

% Tab.~\ref{tab:video_segment} reports the detailed results of these schemes. 
% In Tab.~\ref{tab:video_segment}, we report the detailed results of these schemes. When the merge rate is relatively low (around 50\%), all methods achieve near-lossless performance, demonstrating the substantial temporal redundancy inherent in video tasks. As more tokens are merged, the fixed segmentation strategy exhibits a noticeable drop in accuracy. In contrast, dynamic segmentation methods such as PruneVid~\cite{huang2024prunevid} and FastVID~\cite{shen2025fastvid} achieve better performance while retaining the same or even fewer tokens. In terms of segmentation latency, HoliTom incurs a significant overhead of 35.5ms due to its use of dynamic programming to determine segment boundaries, which introduces an $\mathcal{O}(n^2)$ computational cost. In contrast, other dynamic segmentation methods, such as PruneVid and FastVID, require only 3.5ms and 2.5ms, respectively, making their latency negligible.
In Tab.~\ref{tab:video_segment}, we report the detailed results of these schemes. When the merge rate is relatively low (around 50\%), all methods achieve near-lossless performance, demonstrating the substantial temporal redundancy inherent in video tasks. As more tokens are merged, the fixed segmentation strategy exhibits a noticeable drop in accuracy. In contrast, dynamic segmentation methods such as PruneVid~\cite{huang2024prunevid} and FastVID~\cite{shen2025fastvid} achieve better performance while retaining the same or even fewer tokens. In terms of segmentation latency, HoliTom incurs an additional overhead of 35.5ms largely due to its use of dynamic programming to determine segment boundaries, which introduces an $\mathcal{O}(n^2)$ computational cost. In contrast, other dynamic segmentation methods (e.g., PruneVid, FastVID), require only 3.5ms and 2.5ms, making their latency negligible.

\input{table/video_merge_prune}
\textbf{Second step:} After obtaining high-quality segments, a natural question arises: \textit{How can we eliminate the temporal redundancy of consecutive frames within each segment?} DyCoke~\cite{tao2025dycoke} preserves the visual tokens in the first frame while pruning those in subsequent frames (Fig.~\ref{fig:video_steps} (b)), whereas PruneVid~\cite{huang2024prunevid} merges similar tokens to reduce redundancy (Fig.~\ref{fig:video_steps} (a)). To compare these two schemes, we conduct both temporal merge and prune experiments for DyCoke and PruneVid.

Interestingly, we observe similar results for these two algorithms (Tab.~\ref{tab:video_merge_prune}). Temporal merge outperforms temporal prune (61.22 \emph{vs.} 60.77 for DyCoke and 60.86 \emph{vs.} 60.65 for PruneVid). \textit{These findings indicate that merge may be more suitable than prune for overcoming temporal redundancy.} The main reason is the high similarity between tokens at corresponding positions across consecutive frames within the same segment. See the Appendix for details.

% \subsection{From single-turn to multi-turn dialogue}

% \begin{takeawayv1}[Takeaways of Section 3]
% \textbf{Spatial Redundancy:}
% \begin{itemize}
% \item Vision Tower: Similarity-based and Attention-based metrics perform comparably, while Prune generally outperforms Merge.

% \item LLM: Similarity-based outperforms Attention-based metrics, and Prune remains superior to Merge.
% \end{itemize}

% \textbf{Temporal Redundancy:} 

% Dynamic segmentation outperforms Fixed ones. Merge is slightly more effective than prune.
% \end{takeawayv1}

\findingbox{

\textbf{Spatial Redundancy:} \ding{192} Vision Tower: Similarity-based and Attention-based metrics perform comparably, while Prune generally outperforms Merge. \ding{193} LLM: Sophisticated metrics and solutions should be chosen to suit different scenarios. 

\textbf{Temporal Redundancy:} Dynamic segmentation outperforms Fixed ones. Merge is slightly more effective than prune.
}

% \begin{takeawayv1}[Takeaways of Section 3]
% Spatial Redundancy:

% Vision Tower: Similarity$\approx$Attention, Prune\textgreater Merge.

% LLM: Similarity\textgreater Attention, Prune\textgreater Merge.

% Temporary Redundancy: 

% Dynamic\textgreater Fixed, Prune\textless Merge.

% \end{takeawayv1}

\section{Compression Struggles in Practical Tasks}

% modern inference engine widely use prefixcache in multi-turn 

Although the aforementioned methods~\cite{yang2025visionzip,zhang2024sparsevlm,wen2025stop,alvar2025divprune} have demonstrated strong performance on general single-turn VQA tasks, they neglect systematic evaluation on practical tasks. First, real-world applications often involve fine-grained tasks that require a far more precise understanding of visual details, such as DocVQA~\cite{tito2023hierarchical} and OCRBench~\cite{liu2024ocrbench}. Second, modern inference engines~\cite{zheng2024sglang,gong2025past} for VLMs widely use prefix caching in multi-turn dialogue, reusing encoded visual-text prefixes to reduce redundant computation. Hence, evaluating the effectiveness of compression techniques on both task types is essential.

\input{table/hard_task}
\input{table/multi_turn}
\noindent \textbf{Fine-Grained Tasks.} 
We select two detail-sensitive tasks, DocVQA~\cite{tito2023hierarchical} and OCRBench~\cite{liu2024ocrbench}, to evaluate the accuracy degradation introduced by token reduction under different compression ratios on LLaVA-NeXT-7B~\cite{liu2024llavanext}. \textit{In Tab.~\ref{tab:doc_orc}, we observe that the accuracy of token reduction on these tasks is still far from satisfactory.} For example, retaining approximately 160 tokens leads to non-trivial performance degradation on these tasks, with accuracy dropping to around 50\%, in stark contrast to the around 80\% accuracy achieved on general VQA benchmarks (Tab.~\ref{tab:vision_tower_main} and Tab.~\ref{tab:llm_main}). Therefore, subsequent works should place greater emphasis on these more challenging fine-grained tasks.

% However, many of the aforementioned token reduction methods~\cite{chen2024image,zhang2024sparsevlm,xing2024pyramiddrop,shao2025holitom,liu2024multi} leverage the current-turn question prompt to prune visual tokens and have demonstrated strong performance in single-turn dialogue tasks. 
% such question-dependent approaches may prune visual tokens that are required in subsequent turns, leading to performance degradation in multi-turn dialogue scenarios.
% However, the existing multi-turn dialogue datasets are few and less than satisfactory. They fail to meet the requirements for reliably evaluating VLMs after token reduction. 
% Nevertheless, these methods lack a systematic evaluation on multi-turn dialogue tasks. 

\noindent \textbf{Multi-Turn Tasks.} We further build a multi-turn dialogue dataset tailored for token reduction in a simple-yet-effective manner. Instead of constructing a dataset from scratch, we build upon the existing visual question answering benchmark, GQA~\cite{hudson2019gqa}, by selecting two semantically distinct questions for each image, serving as the first-turn and second-turn questions. Since the first-turn and second-turn questions are different, and their difficulty levels may vary randomly, we further swap the order of the two questions to create a new question pair. As a result, each question appears once in the first turn and once in the second turn, allowing us to eliminate randomness.

Moreover, rather than relying solely on traditional accuracy-based metrics, we focus on evaluating an algorithm's consistency in multi-turn dialogue. Specifically, we are interested in the probability that a question is answered correctly in the second turn, if it can be answered correctly in the first turn. To compute this, we propose a novel metric that measures this conditional accuracy:
\begin{equation}
    P( Q_2^T  | Q_1^T )  = \frac{P(Q_2^T, Q_1^T)}{P(Q_1^T)} \approx \frac{N(Q_2^T, Q_1^T)}{N(Q_1^T)}
    \label{eq:multi_turn}
\end{equation}
where $Q_i^T$ indicates that the question at turn $i$ is answered correctly. $P$ and $N$ denote the probability and corresponding number. We select several representative algorithms and categorize them based on the extent to which they utilize textual information. As we can see in Tab.~\ref{tab:multi_turn}, text-agnostic schemes, such as VisionZip and DivPrune, surpass text-relevant ones (\ie, FastV and SparseVLM). \textit{This suggests that question-dependent approaches may prune visual tokens that are required in subsequent turns, leading to performance degradation in multi-turn dialogue scenarios.} Visualization results can be found in the Appendix.

\findingbox{Token reduction suffers from significant accuracy drops on fine-grained tasks, and prompt-dependent methods yield unsatisfactory results in multi-turn dialogue. Future work should give more attention to the performance in these tasks.}

\section{Achieving Extreme Compression}

\begin{figure}[t]
\begin{center}
     \includegraphics[width=1\linewidth]{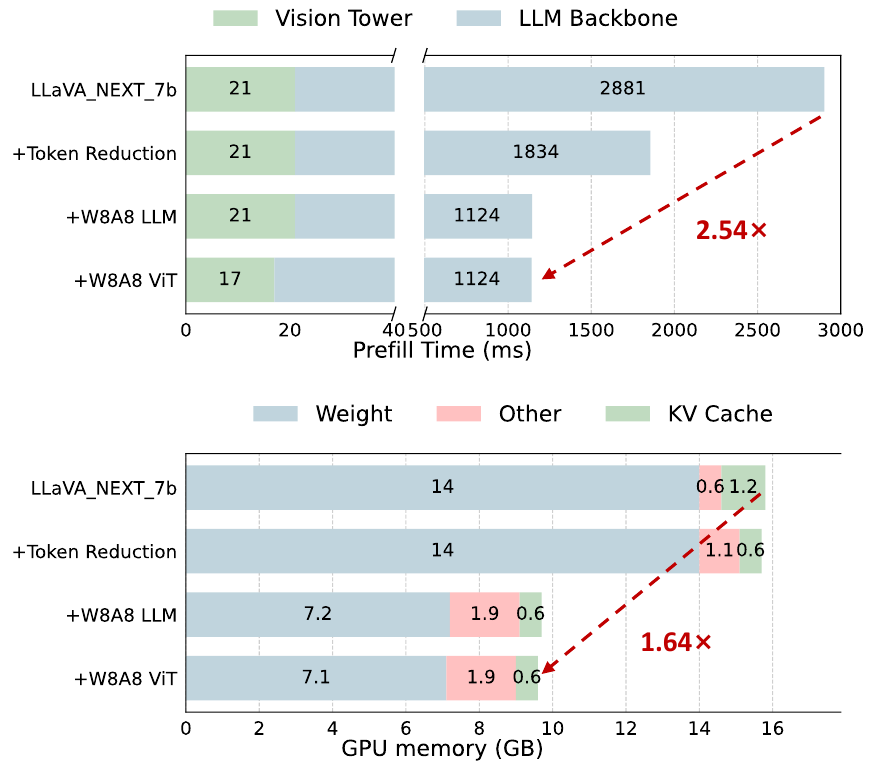}
\end{center}
  \caption{Real inference efficiency on LLaVA-NeXT~\cite{liu2024llavanext}.}
    \label{fig:prefill_memory}
\end{figure}

Although token reduction can significantly reduce inference latency, it does not lower the peak memory usage during inference. The rationale lies in the fact that model weights dominate memory consumption, which often accounts for over 90\%, while token reduction primarily reduces the storage cost of the KV Cache (shown in Fig.~\ref{fig:prefill_memory}). To further address the memory bottleneck in VLM inference, we adopt post-training quantization~\cite{xiao2023smoothquant,frantar2022gptq,lin2024awq}, which is more practical during deployment and application.

\input{table/quant}

Following the categorization of quantization methods, we choose GPTQ~\cite{frantar2022gptq} as a representative algorithm for weight-only quantization (W4A16), and SmoothQuant~\cite{xiao2023smoothquant} for weight-activation quantization (W8A8). In Tab.~\ref{tab:quant}, we present detailed results for the joint application of token reduction and quantization. We summarize the following key observations: 1) Using quantization alone (\eg, W8A8 and W4A16) can retain near-lossless accuracy on VLMs. When combined with token reduction, overall performance mainly depends on the effectiveness of the token reduction method. 2) W8A8 consistently achieves better accuracy than W4A16, regardless of whether token reduction is applied.

To further validate the actual speedup and memory savings of quantized VLM, we deploy LLaVA-NeXT-7B on an NVIDIA RTX 4090 GPU. We utilize the int8 kernel from vLLM~\cite{kwon2023efficient} for ``LLM Backbone" and the int8 kernel from q8~\cite{q8} for ``Vision Tower". When the number of visual tokens decreases from 2487 to 1243 (50\% is considered a safe pruning ratio~\cite{chen2024image}), the model achieves a 1.56$\times$ speedup, but with negligible change in memory usage. However, when combined with quantization, the model achieves a 2.54$\times$ speedup along with a 1.64$\times$ reduction in memory consumption. Therefore, our recommended best practice is to combine token reduction with efficient \textit{model-level} compression techniques, such as quantization.

\findingbox{Introducing relatively stable W8A8 or W4A16 quantization into token reduction yields greater compression rates without a significant performance drop.}

\section{Conclusion}

In this study, we present a VLM compression benchmark with a versatile toolkit, dubbed LLMC+. We identify three major limitations in existing research and conduct comprehensive experiments to evaluate various compression methods. Our benchmark contributes in three key areas: (1) We explore diverse technical strategies for handling spatial and temporal redundancy. For each, we distill the core techniques and perform modular comparisons across related methods, and provide in-depth analysis. (2) Our benchmark systematically evaluates compression performance on fine-grained tasks and multi-turn dialogue, revealing the limitations of existing methods in practical scenarios. (3) We investigate the effectiveness and trade-offs of combining multiple compression methods (\ie, token reduction and quantization). We expect our work to provide valuable insights and serve as a practical guide for the community in advancing more effective VLM compression techniques.

\section{Acknowledgements}
This work was supported by the Beijing Natural Science Foundation (QY24138), the Fundamental Research Funds for the Central Universities, the Postdoctoral Fellowship Program and China Postdoctoral Science Foundation (No. BX20250487).

\bibliography{aaai2026}

\clearpage
\twocolumn[%
\vbox{%
  \hsize\textwidth%
  \linewidth\hsize%
  \vskip 0.625in minus 0.125in%
  \centering%
  {\LARGE\bf Supplementary Material \par}%
  \vskip 1em plus 2fil%
}%
]

\input{supp_content}

\end{document}

%% file: table/classification.tex
\definecolor{forestgreen}{RGB}{34, 139, 34}
\begin{table}[t]
\centering
\resizebox{\linewidth}{!}{
\begin{tabular}{l|l|cc|cc|c}
\toprule
\multirow{2}{*}{Method} & \multirow{2}{*}{Venue} 
& \multicolumn{2}{c|}{Vision} 
& \multicolumn{2}{c|}{Language} & \multirow{2}{*}{Progressive} \\
\cmidrule(lr){3-4} \cmidrule(lr){5-6}
& & Prune & Merge & Prune & Merge &  \\
\midrule
ToMe         & ICLR 2023      & \textcolor{red}{\textbf{\ding{55}}}           & S   & \textcolor{red}{\textbf{\ding{55}}}          & \textcolor{red}{\textbf{\ding{55}}}         & \textcolor{forestgreen}{\textbf{\ding{51}}} \\
FastV        & ECCV 2024      & \textcolor{red}{\textbf{\ding{55}}}           & \textcolor{red}{\textbf{\ding{55}}}             & A   & \textcolor{red}{\textbf{\ding{55}}}         & \textcolor{red}{\textbf{\ding{55}}} \\
SparseVLM    & ICML 2025      & \textcolor{red}{\textbf{\ding{55}}}           & \textcolor{red}{\textbf{\ding{55}}}             & A  & S& \textcolor{forestgreen}{\textbf{\ding{51}}} \\
PDrop        & CVPR 2025      & \textcolor{red}{\textbf{\ding{55}}}           & \textcolor{red}{\textbf{\ding{55}}}             & A    & \textcolor{red}{\textbf{\ding{55}}}         & \textcolor{forestgreen}{\textbf{\ding{51}}} \\
VisionZip    & CVPR 2025      & A      & S       & \textcolor{red}{\textbf{\ding{55}}}          & \textcolor{red}{\textbf{\ding{55}}}         & \textcolor{red}{\textbf{\ding{55}}} \\
VisPruner    & ICCV 2025      & A+S & \textcolor{red}{\textbf{\ding{55}}}         & \textcolor{red}{\textbf{\ding{55}}}          & \textcolor{red}{\textbf{\ding{55}}}         & \textcolor{red}{\textbf{\ding{55}}} \\
DART         & EMNLP 2025  & \textcolor{red}{\textbf{\ding{55}}}           & \textcolor{red}{\textbf{\ding{55}}}             & S  & \textcolor{red}{\textbf{\ding{55}}}         & \textcolor{red}{\textbf{\ding{55}}} \\
DivPrune     & CVPR 2025      & S  & \textcolor{red}{\textbf{\ding{55}}}             & \textcolor{red}{\textbf{\ding{55}}}          & \textcolor{red}{\textbf{\ding{55}}}         & \textcolor{red}{\textbf{\ding{55}}} \\
MustDrop     & arXiv 2024  & A      & S     & A    & \textcolor{red}{\textbf{\ding{55}}}         & \textcolor{red}{\textbf{\ding{55}}} \\
HoliTom      & NeurIPS 2025  & \textcolor{red}{\textbf{\ding{55}}}           & S    & \textcolor{red}{\textbf{\ding{55}}}          & S    & \textcolor{red}{\textbf{\ding{55}}} \\
\bottomrule
\end{tabular}
}
% of token reduction 
\caption{Solutions for spatial redundancy. ``A" and ``S" denote attention- and similarity-based metrics, respectively.}
\label{tab:classify}
\end{table}

\begin{table}[t]
\centering
% \small\setlength
\setlength{\tabcolsep}{4mm}
\resizebox{\linewidth}{!}{
% {\fontsize{9}{11}\selectfont
\begin{tabular}{l|l|ccc}
\toprule
Method & Venue & Segment & Prune & Merge \\
\midrule
DyCoke     & CVPR 2025   & F   & \textcolor{forestgreen}{\textbf{\ding{51}}} & \textcolor{red}{\textbf{\ding{55}}} \\
PruneVid   & ACL 2025    & D & \textcolor{red}{\textbf{\ding{55}}}     & \textcolor{forestgreen}{\textbf{\ding{51}}} \\
FastVID    & NeurIPS 2025  & D      & \textcolor{red}{\textbf{\ding{55}}}     & \textcolor{forestgreen}{\textbf{\ding{51}}} \\
HoliTom    & NeurIPS 2025  & D      & \textcolor{red}{\textbf{\ding{55}}}     & \textcolor{forestgreen}{\textbf{\ding{51}}} \\
\bottomrule
\end{tabular}
}
%of token reduction 
\caption{Solutions for temporal redundancy. ``F" and ``D" denote Fixed and Dynamic segmentation strategies.}
\label{tab:classify_video}
\end{table}

%% file: table/vision_tower.tex
\renewcommand{\multirowsetup}{\centering}

\begin{table*}[t!]
\renewcommand{\arraystretch}{1.1}
\setlength{\tabcolsep}{12pt}
\resizebox{\linewidth}{!}{
\begin{tabular}{llc|cc|cc|cc}
\noalign{\hrule height 1pt}
\textbf{Model} & \textbf{Method} & \textbf{Type} & \textbf{Acc.} & \textbf{Rel.} & \textbf{Acc.} & \textbf{Rel.} & \textbf{Acc.} & \textbf{Rel.} \\
\noalign{\hrule height 1pt}
\rowcolor{gray!20}
% LLaVA-1.5-7B~\cite{liu2024improved}
\multicolumn{3}{c}{} & \multicolumn{6}{c}{\textit{Upper Bound, 576 Tokens (100\%)}} \\
& Vanilla & - & 64.3 & 100\% & 64.3 & 100\% & 64.3 & 100\% \\
\noalign{\hrule height 1pt}
\rowcolor{gray!20}
\multicolumn{3}{c}{} & \multicolumn{2}{c}{\textit{192 Tokens ($\downarrow$ 66.7\%)}} &  \multicolumn{2}{c}{\textit{128 Tokens ($\downarrow$ 77.8\%)}} & \multicolumn{2}{c}{\textit{64 Tokens ($\downarrow$ 88.9\%)}}\\
\multirow{7}{*}{LLaVA-1.5-7B~\cite{liu2024improved}}& VisionZip~\cite{yang2025visionzip} & PA & 62.9 & 97.9\% & 62.1 & 96.7\% & 60.5 & 94.1\% \\
& VisPruner~\cite{zhang2024beyond} & PS & 62.5 & 97.2\% & 61.3 & 95.4\% & 59.7 & 92.8\% \\
& DivPrune~\cite{alvar2025divprune} & PS & 62.9 & 97.8\% & 62.4 & 97.1\% & 60.9 & 94.7\% \\
& ToMe~\cite{bolya2022token} & MS & 61.6 & 95.8\% & 61.0 & 94.9\% & 59.2 & 92.1\% \\
& VisionZip~\cite{yang2025visionzip} & MS & 60.8 & 94.5\% & 59.4 & 92.4\% & 56.3 & 87.5\% \\
& MustDrop~\cite{liu2024multi} & MS & 61.7 & 95.9\% & 59.3 & 92.3\% & 52.0 & 80.9\% \\
& VisionZip~\cite{yang2025visionzip} & PA+MS & 62.9 & 97.9\% & 62.2 & 96.7\% & 60.4 & 93.9\% \\
% \midrule
% \midrule
\noalign{\hrule height 1pt}
% \noalign{\hrule height 1pt}
\rowcolor{gray!20}
\multicolumn{3}{c}{} & \multicolumn{6}{c}{\textit{Upper Bound, $\sim$2880 Tokens (100\%)}} \\
 & Vanilla & - & 68.6 & 100\% & 68.6 & 100\% & 68.6 & 100\% \\
\noalign{\hrule height 1pt}
\rowcolor{gray!20}
\multicolumn{3}{c}{} & \multicolumn{2}{c}{\textit{$\sim$640 Tokens ($\downarrow$ 66.7\%)}} &  \multicolumn{2}{c}{\textit{$\sim$320 Tokens ($\downarrow$ 77.8\%)}} & \multicolumn{2}{c}{\textit{$\sim$160 Tokens ($\downarrow$ 88.9\%)}}\\
\multirow{5}{*}{LLaVA-NeXT-7B~\cite{liu2024llavanext}}&VisionZip~\cite{yang2025visionzip} & PA & 66.8 & 97.3\% &  64.9 & 94.6\%  &  62.2 & 90.6\%  \\
&VisPruner~\cite{zhang2024beyond} & PS & 64.9 & 94.6\% &  62.7 & 91.4\%  &  59.6 & 86.8\%  \\
&DivPrune~\cite{alvar2025divprune} & PS & 65.6 & 95.5\% &  63.8 & 93.0\%  &  62.3 & 90.8\%  \\
&VisionZip~\cite{yang2025visionzip} & MS & 62.7 & 91.4\% &  60.1 & 87.6\%  &  56.1 & 81.8\%  \\
&MustDrop~\cite{liu2024multi} & MS & 64.9 & 94.5\% &  61.3 & 89.2\%  & - & - \\

\noalign{\hrule height 1pt}
\rowcolor{gray!20}
\multicolumn{3}{c}{} & \multicolumn{6}{c}{\textit{Upper Bound, 100\% Tokens}} \\
&Vanilla & -&79.3 & 100\% & 79.3 & 100\% & 79.3 & 100\% \\
\noalign{\hrule height 1pt}
\rowcolor{gray!20}
\multicolumn{3}{c}{} & \multicolumn{2}{c}{\textit{33.3\% Tokens ($\downarrow$ 66.7\%)}} &  \multicolumn{2}{c}{\textit{22.2\% Tokens ($\downarrow$ 77.8\%)}} & \multicolumn{2}{c}{\textit{11.1\% Tokens ($\downarrow$ 88.9\%)}}\\
\multirow{4}{*}{Qwen2.5-VL-7B~\cite{bai2025qwen2}}&VisionZip~\cite{yang2025visionzip} & PA & 77.9 & 98.2\% & 75.9 & 95.7\% & 70.7 & 89.2\% \\
&VisPruner~\cite{zhang2024beyond} & PS & 76.8 & 96.8\% & 74.4 & 93.9\% & 68.8 & 86.7\% \\
&DivPrune~\cite{alvar2025divprune} & PS & 77.3 & 97.4\% & 75.6 & 95.3\% & 71.5 & 90.2\% \\
&VisionZip~\cite{yang2025visionzip} & MS & 76.9 & 97.0\% & 74.6 & 94.1\% & 68.3 & 86.1\% \\
\noalign{\hrule height 1pt}
\end{tabular}
}
\caption{Average performance of different token reduction methods in Vision Tower across seven benchmarks\footnotemark.}
\label{tab:vision_tower_main}
\end{table*}

\footnotetext{The seven benchmarks include GQA, MMB\_EN, MME, POPE, TextVQA, VizWiz\_VQA, and ScienceQA. Detailed results are provided in the \textbf{Appendix}.}
% Due to space constraints, we report only the average accuracy and average percentage scores in the main text; detailed results are provided in the appendix.

\begin{table}[t]
\renewcommand{\arraystretch}{1.1}

% \vspace{-5pt}
% \setlength{\tabcolsep}{5pt}
\resizebox{\linewidth}{!}{
\begin{tabular}{lc|cc|c}
\noalign{\hrule height 1pt}
\textbf{Method} & \textbf{Type} & \textbf{Acc.} & \textbf{Rel.} &  \textbf{Prefill}\\
\noalign{\hrule height 1pt}
\rowcolor{gray!20}

\multicolumn{5}{c}{\textit{Upper Bound, 576 Tokens (100\%)}}\\
LLaVA-1.5-7B~\cite{liu2024improved} & - & 64.3 & 100\% & 27.3 ms\\
\noalign{\hrule height 1pt}
\rowcolor{gray!20}
\multicolumn{5}{c}{\textit{Retain 192 Tokens in Average ($\downarrow$ 66.7\%)}} \\
% ToMe~\cite{bolya2022token} & MS & 54.3 & 60.5 & -\\
ToMe\textsuperscript{*}~\cite{bolya2022token}
 & MS & 53.5 & 83.2\% & 16.6 ms \\
ToMe\textsuperscript{\dag}~\cite{bolya2022token} & MS & 62.4 & 97.0\% & 16.6 ms \\
MustDrop\textsuperscript{*}~\cite{liu2024multi} & MS & 58.8 & 91.4\% & 16.6 ms \\
MustDrop\textsuperscript{\dag}~\cite{liu2024multi} & MS & 61.7 & 95.9\% & 16.6 ms \\
\noalign{\hrule height 1pt}
\rowcolor{gray!20}
\multicolumn{5}{c}{\textit{Retain 128 Tokens in Average ($\downarrow$ 77.8\%)}} \\
ToMe\textsuperscript{*}~\cite{bolya2022token} & MS & 45.3 & 70.5\% & 15.8 ms \\
ToMe\textsuperscript{\dag}~\cite{bolya2022token} & MS & 60.3 & 93.8\% & 16.0 ms \\
MustDrop\textsuperscript{*}~\cite{liu2024multi} & MS & 54.2 & 84.3\% & 16.0 ms \\
MustDrop\textsuperscript{\dag}~\cite{liu2024multi} & MS & 59.3 & 92.3\% & 16.2 ms \\
\noalign{\hrule height 1pt}
\end{tabular}
}

\caption{Results of token reduction at different layers in Vision Tower. \textsuperscript{*} and \textsuperscript{\dag} denote merging in the first and last layer.}
\label{tab:vision_tower_pos}
\end{table}

%% file: table/llm.tex
\renewcommand{\multirowsetup}{\centering}

\begin{table*}[t!]
\renewcommand{\arraystretch}{1.1}
\setlength{\tabcolsep}{12pt}
\resizebox{\linewidth}{!}{
\begin{tabular}{llc|cc|cc|cc}
\noalign{\hrule height 1pt}
\textbf{Model} & \textbf{Method} & \textbf{Type} & \textbf{Acc.} & \textbf{Rel.} & \textbf{Acc.} & \textbf{Rel.} & \textbf{Acc.} & \textbf{Rel.} \\
\noalign{\hrule height 1pt}
\rowcolor{gray!20}
% LLaVA-1.5-7B~\cite{liu2024improved}
\multicolumn{3}{c}{} & \multicolumn{6}{c}{\textit{Upper Bound, 576 Tokens (100\%)}} \\
& Vanilla & - & 64.3 & 100\% & 64.3 & 100\% & 64.3 & 100\% \\
\noalign{\hrule height 1pt}
\rowcolor{gray!20}
\multicolumn{3}{c}{} & \multicolumn{2}{c}{\textit{192 Tokens ($\downarrow$ 66.7\%)}} &  \multicolumn{2}{c}{\textit{128 Tokens ($\downarrow$ 77.8\%)}} & \multicolumn{2}{c}{\textit{64 Tokens ($\downarrow$ 88.9\%)}}\\
\multirow{5}{*}{LLaVA-1.5-7B~\cite{liu2024improved}}& FastV~\cite{chen2024image} & PA & 62.4 & 97.1\% & 60.9 & 94.7\% & 56.8 & 88.3\% \\
& SparseVLM~\cite{zhang2024sparsevlm} & PA & 63.3 & 98.5\% & 62.4 & 97.1\% & 60.0 & 93.3\% \\
& DART~\cite{wen2025stop} & PS & 63.7 & 99.0\% & 62.9 & 97.9\% & 61.4 & 95.5\% \\
& HoliTom~\cite{shao2025holitom} & MS & 61.5 & 95.6\% & 60.1 & 93.5\% & 57.5 & 89.4\% \\
& SparseVLM~\cite{zhang2024sparsevlm} & PA+MS & 63.2 & 98.2\% & 62.6 & 97.3\%  & 60.5 & 94.0\%  \\

\noalign{\hrule height 1pt}
% \noalign{\hrule height 1pt}
\rowcolor{gray!20}
\multicolumn{3}{c}{} & \multicolumn{6}{c}{\textit{Upper Bound, $\sim$2880 Tokens (100\%)}} \\
 & Vanilla & - & 68.6 & 100\% & 68.6 & 100\% & 68.6 & 100\% \\
\noalign{\hrule height 1pt}
\rowcolor{gray!20}
\multicolumn{3}{c}{} & \multicolumn{2}{c}{\textit{$\sim$640 Tokens ($\downarrow$ 66.7\%)}} &  \multicolumn{2}{c}{\textit{$\sim$320 Tokens ($\downarrow$ 77.8\%)}} & \multicolumn{2}{c}{\textit{$\sim$160 Tokens ($\downarrow$ 88.9\%)}}\\
\multirow{4}{*}{LLaVA-NeXT-7B~\cite{liu2024llavanext}}&FastV~\cite{chen2024image} & PA & 65.8 & 95.9\% & 61.7 & 89.9\% & 56.0 & 81.5\% \\
&SparseVLM~\cite{zhang2024sparsevlm} & PA & 66.9 & 97.5\% & 64.8 & 94.5\% & 61.3 & 89.2\% \\
&DART~\cite{wen2025stop} & PS & 67.2 & 98.0\% & 65.2 & 95.0\% & 62.1 & 90.5\% \\
&HoliTom~\cite{shao2025holitom} & MS & 64.7 & 94.2\% & 61.5 & 89.6\% & 54.6 & 79.6\% \\
\noalign{\hrule height 1pt}
\rowcolor{gray!20}
\multicolumn{3}{c}{} & \multicolumn{6}{c}{\textit{Upper Bound, 100\% Tokens}} \\
&Vanilla & -&79.3 & 100\% & 79.3 & 100\% & 79.3 & 100\% \\
\noalign{\hrule height 1pt}
\rowcolor{gray!20}
\multicolumn{3}{c}{} & \multicolumn{2}{c}{\textit{33.3\% Tokens ($\downarrow$ 66.7\%)}} &  \multicolumn{2}{c}{\textit{22.2\% Tokens ($\downarrow$ 77.8\%)}} & \multicolumn{2}{c}{\textit{11.1\% Tokens ($\downarrow$ 88.9\%)}}\\
\multirow{4}{*}{Qwen2.5-VL-7B~\cite{bai2025qwen2}}& FastV~\cite{chen2024image} & PA & 75.9 & 95.7\% & 72.5 & 91.4\% & 64.5 & 81.3\% \\
&SparseVLM~\cite{zhang2024sparsevlm} & PA & 77.6 & 97.9\% & 76.1 & 95.9\% & 72.0 & 90.7\% \\
&DART~\cite{wen2025stop} & PS & 76.6 & 96.6\% & 74.6 & 94.1\% & 69.2 & 87.3\% \\
&HoliTom~\cite{shao2025holitom} & MS & 77.2 & 97.4\% & 75.0 & 94.6\% & 69.8 & 88.1\% \\
\noalign{\hrule height 1pt}
\end{tabular}
}
\caption{Average performance of different token reduction methods in LLM across seven benchmarks.}
\label{tab:llm_main}
\end{table*}

%% file: table/segment.tex
\begin{table*}[t]
\centering
\setlength{\tabcolsep}{8pt}

\resizebox{\linewidth}{!}{
\begin{tabular}{c|cccc|cccc|cc}
\toprule
Method & 
MR&
RR& 
Segment&
Prefill& 
MVBench & 
LongVideoBench & 
MLVU & 
VideoMME &
% VideoMME
Score & \% \\
% \cline{7-10}
% \cmidrule(lr){7-10}
% & & & & & &  &  &  \\
\midrule
Duration & && && 16 sec & 1$\sim$60 min & 3$\sim$120 min & 1$\sim$60 min &  & \\
\midrule
Vanilla & - &100\% & - & 290.7ms & 58.35 & 56.55 & 63.10 & 58.51 & 59.13 & 100.0\%  \\
\midrule
% \textbf{merge rate=50\%} & & & &  &  &  &  &  \\
Fixed       & 50\%&56.2\% & 0 ms     & 164.7 ms & 58.67 & 55.87 & 63.65 & 59.22 & 59.35 & 100.4\% \\
% Fixed k=8       & 62.5\% & 0 ms     & 178.7 ms & 58.72 & 56.47 & 62.45 & 59.37 & 59.25 & 100.2\% \\
PruneVid    & 50\%&54.6\% & 3.3 ms   & 159.5 ms & 58.45 & 55.80 & 62.76 & 58.70 & 58.93 & 99.7\% \\
% PruneVid k=8    & 56.5\% & 3.5 ms   & 164.1 ms & 58.22 & 55.27 & 62.91 & 58.25 & 58.66 & 99.2\% \\
FastVID     & 50\%&56.2\% & 2.5 ms   & 165.7 ms & 58.35 & 54.90 & 62.98 & 59.00 & 58.81 & 99.5\% \\
% FastVID k=8     & 62.5\% & 2.5 ms   & 180.0 ms & 58.10 & 55.87 & 63.21 & 58.77 & 58.99 & 99.7\%\\
\midrule

Fixed & 80\%&30.3\% & 0 ms   & 88.1 ms  & 57.98 & 54.37 & 60.79 & 58.18 & 57.83 & 97.8\% \\
% Fixed k=8 & 40.3\% & 0 ms   & 122.4 ms & 58.52 & 56.10 & 61.79 & 58.44 & 58.71 & 99.3\% \\
PruneVid &80\%& 27.7\% & 3.3 ms & 82.5 ms  & 57.80 & 54.53 & 61.89 & 57.34 & 57.89 & 97.9\% \\
% PruneVid k=8 & 30.9\% & 3.5 ms & 90.7 ms  & 58.13 & 54.67 & 61.51 & 57.70 & 58.00 & 98.1\% \\
FastVID &80\%& 30.3\% & 2.5 ms & 88.1 ms  & 57.90 & 54.82 & 61.69 & 57.81 & 58.06 & 98.2\% \\
% FastVID k=8 & 40.3\% & 2.5 ms & 119.2 ms & 58.03 & 55.87 & 63.50 & 58.33 & 58.93 & 99.7\%\\

\midrule
HoliTom     & - &52.4\% & 35.5 ms  & 149.0 ms & 57.70 & 56.17 & 63.42 & 59.33 & 59.16 & 100.1\% \\
HoliTom     & - & 68.9\% & 35.5 ms  & 199.6 ms & 58.35 & 55.80 & 63.30 & 58.85 & 59.08 & 99.9\% \\

\bottomrule
\end{tabular}
}

\caption{Comparison of four video segmentation methods on LLaVA-OneVision~\cite{li2024llava}. \textbf{M}erge \textbf{R}ate (MR) denotes the intra-segment merge ratio, whereas \textbf{R}etention \textbf{R}ate (RR) measures the overall token retention for an input. This also implies that even with the same Merge Rate, different segment lengths will lead to varying Retention Rates.
% MR and RR denote Merge Rate and Retention Rate, respectively.
}
\label{tab:video_segment}
\end{table*}

%% file: table/video_merge_prune.tex
\begin{table}[t]
\centering
\setlength{\tabcolsep}{1mm}
\resizebox{\linewidth}{!}{
\begin{tabular}{c|cccc|c}
\toprule
Method & 
MVBench & 
LongVideoBench & 
MLVU & 
VideoMME &
Avg.  \\
\midrule
Duration & 16 sec & 1$\sim$60 min & 3$\sim$120 min & 1$\sim$60 min &   \\
\midrule
DyCoke$^{*}$     & 58.35 & 63.10 & 63.58 & 59.44 & 61.12 \\
DyCoke$^{\dag}$     & 58.23 & 63.10 & 62.99 & 58.74 & 60.77 \\
PruneVid$^{*}$   & 57.28 & 63.30 & 63.30 & 59.55 & 60.86 \\
PruneVid$^{\dag}$   & 56.58 & 63.31 & 63.31 & 59.40 & 60.65 \\

\bottomrule
\end{tabular}
}

\caption{Comparison between Temporal Merge (superscript $*$) and Temporal Prune (superscript $\dag$).}
\label{tab:video_merge_prune}
\end{table}

%% file: table/hard_task.tex
\begin{table}[t]
\centering
\setlength{\tabcolsep}{3mm}
\resizebox{\linewidth}{!}{
\begin{tabular}{l|cc|cc|cc}
\toprule
\multirow{2}{*}{Method} & 
\multicolumn{2}{c|}{$\sim$640 Tokens} & 
\multicolumn{2}{c|}{$\sim$320 Tokens} & 
\multicolumn{2}{c}{$\sim$160 Tokens} \\
\cmidrule(lr){2-3} \cmidrule(lr){4-5} \cmidrule(lr){6-7}
& DOC & OCR & DOC & OCR & DOC & OCR \\
\midrule
Vanilla     & 68.5 & 52.1 & 68.5 & 52.1 & 68.5 & 52.1 \\
\midrule
FastV       & 46.6 & 38.9 & 29.7 & 24.8 & 17.8 & 15.4 \\
DART        & 54.6 & 44.2 & 42.3 & 35.4 & 29.4 & 26.8 \\
SparseVLM   & 49.1 & 40.6 & 31.1 & 29.0 & 17.3 & 18.6 \\
VisionZip   & 56.9 & 48.5 & 42.5 & 39.5 & 28.8 & 29.5 \\
VisPruner   & 53.9 & 44.8 & 39.6 & 37.3 & 27.4 & 31.7 \\
DivPrune    & 44.8 & 37.6 & 32.8 & 33.0 & 26.1 & 26.6 \\
\bottomrule
\end{tabular}
}
\caption{DocVQA~\cite{tito2023hierarchical} and OCRBench~\cite{liu2024ocrbench} results of token reduction.}
\label{tab:doc_orc}
\end{table}

%% file: table/multi_turn.tex
\begin{table}[t]
\centering
\setlength{\tabcolsep}{4mm}
\resizebox{\linewidth}{!}{
\begin{tabular}{l|c|c}
\toprule
Method & Text Info  & Score \\
\midrule
VisionZip~\cite{yang2025visionzip}   & None  &  0.938 \\
DivPrune~\cite{alvar2025divprune}    & None  &  0.929 \\
DART~\cite{wen2025stop}        & Low   &  0.903 \\
FastV~\cite{chen2024image}       & High  & 0.896 \\
SparseVLM~\cite{zhang2024sparsevlm}   & High  &  0.882 \\
\bottomrule
\end{tabular}
}
\caption{Performance comparison of various methods on multi-turn dialogue tasks.}
\label{tab:multi_turn}
\end{table}

%% file: table/quant.tex
\begin{table}[t]
\centering
\resizebox{\linewidth}{!}{
\begin{tabular}{l|c|c|c|c|c|c}
\toprule
Method & GQA & MMB\_EN & MME & POPE & TextVQA & Avg (\%) \\
\midrule
LLaVA-1.5-7B        & 61.2 & 62.9 & 1805 & 85.5 & 48.6 & 100.0 \\
\midrule
FastV                         & 56.1 & 61.8 & 1744 & 79.5 & 42.8 & 93.4 \\
GPTQ                          & 60.9 & 62.8 & 1790 & 85.1 & 48.3 & 99.5 \\
GPTQ + FastV                  & 55.3 & 60.5 & 1732 & 79.2 & 42.6 & 92.5 \\
SQ                            & 61.2 & 63.7 & 1781 & 85.4 & 48.5 & 99.9 \\
SQ + FastV                    & 56.0 & 61.2 & 1702 & 79.6 & 42.9 & 92.9 \\
\midrule
LLaVA-1.5-13B        & 62.6 & 68.3 & 1854 & 85.7 & 52.8 & 100.0 \\
\midrule
FastV                         & 58.6 & 66.8 & 1747 & 80.7 & 45.1 & 93.0 \\
GPTQ                          & 62.2 & 67.4 & 1840 & 85.8 & 52.6 & 99.4 \\
GPTQ + FastV                  & 58.5 & 65.6 & 1749 & 80.8 & 44.9 & 92.6 \\
SQ                            & 62.6 & 68.5 & 1840 & 85.6 & 53.0 & 100.0 \\
SQ + FastV                    & 58.3 & 66.1 & 1757 & 80.6 & 44.9 & 92.8 \\
\bottomrule
\end{tabular}
}
\caption{Results of joint token reduction and quantization on LLaVA-1.5, where SQ denotes SmoothQuant and 192 visual tokens are preserved for FastV.}
\label{tab:quant}
\end{table}

%% file: supp_content.tex
\appendix

% \maketitle
\setcounter{secnumdepth}{2}
\section{Related Works}
\subsection{Efficient Vision-Language Models}
Recent Vision-Language Models (VLMs), such as Flamingo~\cite{alayrac2022flamingo}, BLIP-2~\cite{li2023blip}, MiniGPT-4~\cite{zhu2023minigpt}, and Qwen-VL~\cite{wang2024qwen2}, typically follow an encoder-projector-decoder architecture, where visual inputs are encoded into tokens, aligned via a projector (\eg, MLP or Q-Former), and decoded by a language model. To improve visual understanding, these models adopt inputs with high spatial and temporal resolution, resulting in a surge of visual tokens and increased computation. To address this, several efficient strategies have been proposed: 1) BLIP-2~\cite{li2023blip} introduces Q-Former, which reduces the token count via learned queries. 2) Downsampling methods like bilinear interpolation (LLaVA-OneVision~\cite{li2024llava}) and average pooling (LLaVA-Video~\cite{zhang2024video}) compress token grids. 3) Pixel unshuffle, used in InternVL~\cite{chen2024internvl}, decreases spatial tokens by rearranging feature maps. Despite these advances, their substantial model size and computational overhead pose significant challenges for deployment in resource-constrained devices, highlighting the urgent need for more efficient strategies.

\subsection{Training-free Compression}
Compared with time-intensive and energy-intensive compression methods such as knowledge distillation~\cite{gou2021knowledge}, training-free compression offers a more efficient solution for reducing the cost of VLMs. A primary approach is token compression~\cite{chen2024image,zhang2024sparsevlm,bolya2022token,yang2025visionzip,shi2025static,alvar2025divprune,liu2024multi,shang2024llava,arif2025hired,song2024less,xing2024pyramiddrop,han2024filter,lan2024vidcompress,ye2025atp,vasu2025fastvlm,han2025adafv,yang2025topv,dhouib2025pact,liu2025video,tao2025dycoke,fu2024framefusion,shen2025fastvid,huang2024prunevid,shao2025holitom}, which reduces the number of visual tokens to accelerate inference. These schemes either utilize attention-based~\cite{chen2024image,xing2024pyramiddrop,yang2025visionzip,liu2024multi} or similarity-based~\cite{bolya2022token,alvar2025divprune,zhang2024beyond} metrics to discard or merge less salient tokens. Recently, several methods~\cite{tao2025dycoke,fu2024framefusion,shen2025fastvid,shao2025holitom,shi2025static} have also been proposed specifically for videos, aiming to reduce temporal redundancy across consecutive frames. Another direction is model compression, including quantization~\cite{xiao2023smoothquant,frantar2022gptq,lin2024awq,chen2024db,lv2024ptq4sam,tian2024qvd,huang2024tfmqdmtemporalfeaturemaintenance,huang2025temporalfeaturemattersframework,huang2025qvgenpushinglimitquantized}, network pruning~\cite{ma2023llm,frantar2023sparsegpt,sun2023simple,wnag2024ptsbenchcomprehensiveposttrainingsparsity,huang2025harmonicaharmonizingtraininginference,guo2020channel,guo2020model,guo2020multi,guo2021jointpruning,guo2023multidimensional,he2025da-kd} and low-rank factorization~\cite{zhao2024galore,swaminathan2020sparse}. Among them, quantization stands out as a highly effective and promising approach, offering better performance and compression rate. In this work, we study post-training quantization to complement token reduction and address the memory bottleneck of VLMs.

\section{Implementation Details}

This section outlines the implementation details for each algorithm. By default, token reduction is applied at the output of the Vision Tower unless otherwise stated in Tab. 4, Tab. 6 and Tab. 7.

\input{appendix/vision_tower}

\input{appendix/vision_tower_position}
\input{appendix/llm}

\begin{itemize}
    \item MustDrop~\cite{liu2024multi}: MustDrop MS~\cite{liu2024multi} applies a 3$\times$3 window for spatial merge in Tab. 3 and Tab. 4.
    \item ToMe~\cite{bolya2022token}: ToMe MS distributes the pruning of tokens evenly across all layers in Tab. 3. In Tab. 4, since ToMe can merge at most 50\% of tokens within each layer, we merge the first/last two layers down to 192 tokens, and first/last three layers down to 128 tokens.
    \item VisionZip~\cite{yang2025visionzip}: For the VisionZip PA+MS, we adhere to the Dominant and Contextual numbers reported in the original publication.
    \item SparseVLM~\cite{zhang2024sparsevlm}: For the SparseVLM PA+MS, we follow the token number of Relevant Text Token Selection and Token Recycling in its paper.
    \item FastVID~\cite{shen2025fastvid}: For fair comparison, we remove the segmentation for transitions where the similarity falls below a fixed threshold $\tau$ in Tab. 6.
    \item HoliTom~\cite{shao2025holitom}: We set T to 0.8 and achieved a retention rate of 68.9\%, while setting T to 0.6 resulted in a retention rate of 52.4\% in Tab. 6.
    \item PruneVid~\cite{huang2024prunevid}: The threshold for token reduction in Tab. 7 is set to 0.6.
    \item DyCoke~\cite{tao2025dycoke}: The token reduction rate is set to 0.5 in Tab. 7.
        
\end{itemize}

\section{Detailed Results}

We provide detailed results (Tab.~\ref{tab:supp_vision_llava_7b}, Tab.~\ref{tab:supp_vision_llava_next_7b}, Tab.~\ref{tab:supp_vision_qwen_7b}, Tab.~\ref{tab:supp_vision_tower_pos_llava_7b}, Tab.~\ref{tab:supp_llm_llava_7b}, Tab.~\ref{tab:supp_llm_llava_next_7b} and Tab.~\ref{tab:supp_llm_qwen_7b}) on each dataset as a supplement to the tables (\ie, Tab. 3, Tab. 4 and Tab. 5) in the main paper. Additionally, Tab.~\ref{tab:token_performance} reports the measured prefill and decode latencies, as well as the peak memory usage during runtime on . For both FastV~\cite{chen2024image} and DART~\cite{wen2025stop}, the token reduction is applied at the 5-th block of the LLM. Two key observations can be drawn from the results: (1) Token reduction primarily reduces the prefill latency, while its impact on decode latency and peak memory is negligible. (2) As the model size increases, \eg, from LLaVA-1.5-7B to LLaVA-1.5-13B, the acceleration effect on the prefill stage becomes more pronounced when maintaining the same proportion of retained visual tokens ($\sim2.0\times$ vs. $\sim1.5\times$).

\section{Visualization}

\begin{figure*}[!t]
\begin{center}
     \includegraphics[width=1\linewidth]{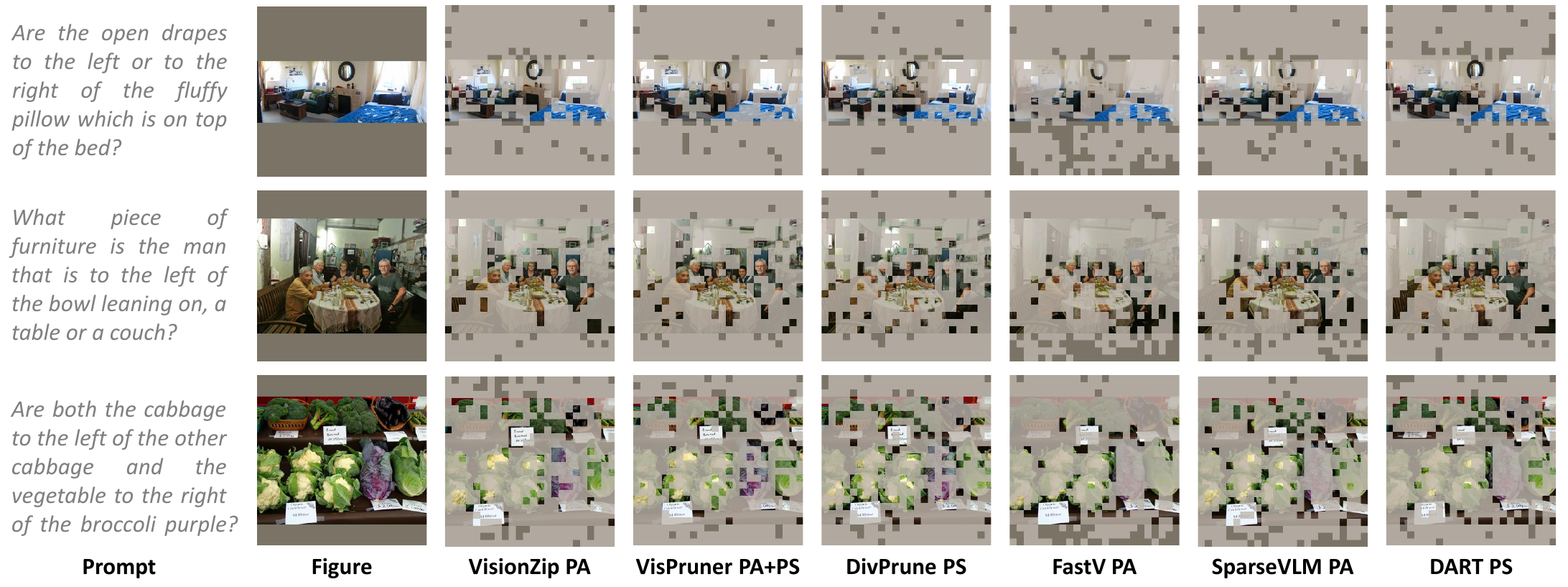}
\end{center}
  \caption{Qualitative results of GQA~\cite{hudson2019gqa} benchmark on LLaVA-1.5-7B~\cite{liu2023visual}.}
    \label{fig:visual_prune}
\end{figure*}

\subsection{Visualization Examples}
In Fig.~\ref{fig:visual_prune}, we present qualitative results of different algorithms on the GQA~\cite{hudson2019gqa} dataset. We observe that text-guided strategies, particularly FastV~\cite{chen2024image} and SparseVLM~\cite{zhang2024sparsevlm}, tend to retain a significant amount of irrelevant background. For example, the black background near the text region, which is located at the bottom of the image, is often preserved. This also explains why DART~\cite{wen2025stop} achieves better accuracy compared to the other two algorithms, as it retains only a very small number of pivot text-relevant tokens and therefore preserves relatively less background information. Therefore, in LLMs, relying entirely on the question prompt to guide token pruning can lead to unsatisfactory results.

In comparison, pruning methods~\cite{yang2025visionzip,alvar2025divprune} in Vision Tower lack access to textual information, which leads to reduced attention to the question prompt. This can be observed in the second image, where objects such as ``table" or ``couch" receive less focus.

\begin{figure*}[!t]
\begin{center}
     \includegraphics[width=1\linewidth]{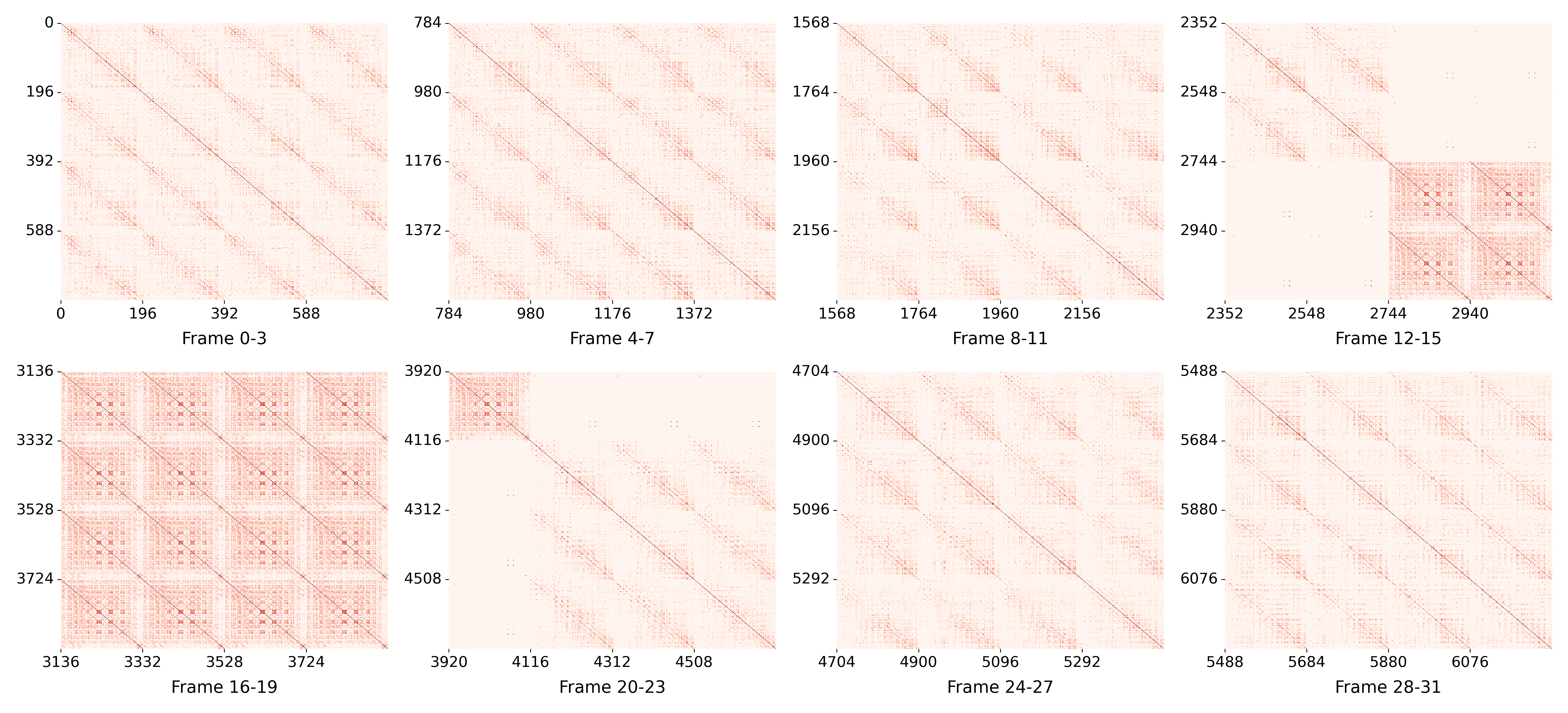}
\end{center}
  \caption{Visual Token similarities in LLaVA-OneVision~\cite{li2024llava}.}
    \label{fig:video_sim}
\end{figure*}
\subsection{Similarity in Video}
We utilize cosine similarity to analyze the redundancy between visual tokens. As shown in Fig.~\ref{fig:video_sim}, LLaVA-OneVision\cite{li2024llava} encodes the video into 32 frames, each represented by 196 feature tokens. We group the tokens from four frames into a single subplot, resulting in a total of eight subplots. To enhance clarity, we visualize only the top 10\% of token similarities. Two types of redundancy mentioned in the main paper (\ie, spatial and temporal redundancy), can be clearly observed from the figure. 

For spatial redundancy, we observe from the cosine similarities in frames 16-19 (\ie, subplot 5) that each frame contains a high degree of intra-frame similarity. In addition, temporal redundancy is primarily reflected in the same spatial positions across different frames, specifically between token indices at positions $idx$ and $idx+196n$, where n is an integer.

\begin{figure*}[!t]
\begin{center}
     \includegraphics[width=1\linewidth]{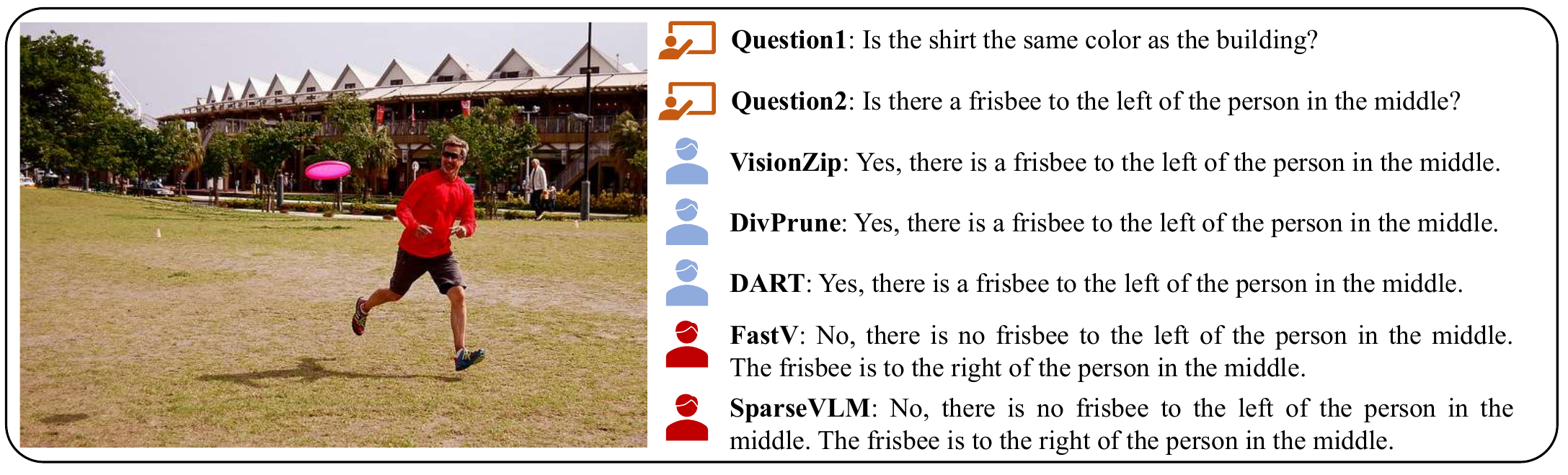}
\end{center}
  \caption{Visualization of multi-turn dialogues across different algorithms.}
    \label{fig:mult-_turn_vis}
\end{figure*}
\subsection{Multi-Turn Dialogue Task}
Fig.~\ref{fig:mult-_turn_vis} shows the visualization of multi-turn dialogues. Specifically, we perform visual token pruning using each algorithm in the first turn and examine their responses to the question in the second turn. It is worth noting that if Question 2 is asked in the first turn, all five algorithms generate the same response: ``Yes, there is a frisbee to the left of the person in the middle." However, when the same question is asked in the second turn, the two text-relevant methods (\ie, FastV and SparseVLM) produce unsatisfactory responses. This further demonstrates that these text-guided token reduction methods are highly dependent on the current-turn question, which may affect the model's ability to answer questions in subsequent turns.

%% file: appendix/vision_tower.tex
\renewcommand{\multirowsetup}{\centering}

\begin{table*}[t!]
\renewcommand{\arraystretch}{1.2}
\resizebox{\linewidth}{!}{
\begin{tabular}{lc|cccccccc}
\noalign{\hrule height 1pt}
\textbf{Method} & \textbf{Type} & \textbf{GQA} & \textbf{MMB} & \textbf{MME} & \textbf{POPE} & \textbf{TextVQA} & \textbf{VizWiz} & \textbf{ScienceQA} & \textbf{OCRBench} \\
\noalign{\hrule height 1pt}
\rowcolor{gray!20}

\multicolumn{10}{c}{\textit{Upper Bound, 576 Tokens (100\%)}}\\
LLaVA-1.5-7B~\cite{liu2023visual} & - & 62.0 & 64.2 & 67.0 & 87.0 & 46.1 & 54.3 & 69.5 & 31.3 \\
\noalign{\hrule height 1pt}
\rowcolor{gray!20}
\multicolumn{10}{c}{\textit{Retain 192 Tokens in Average ($\downarrow$ 66.7\%)}} \\
% ToMe~\cite{bolya2022token} & MS & 54.3 & 60.5 & -\\
VisionZip~\cite{yang2025visionzip} & PA & 59.3 & 63.5 & 63.8 & 86.4 & 44.8 & 54.3 & 68.6 & 31.3 \\
VisPruner~\cite{zhang2024beyond} & PS & 59.6 & 62.5 & 62.2 & 87.6 & 41.7 & 55.4 & 68.3 & 29.2 \\
DivPrune~\cite{alvar2025divprune} & PS & 59.9 & 62.4 & 62.6 & 87.5 & 43.3 & 55.7 & 68.7 & 30.0 \\
ToMe~\cite{bolya2022token} & MS & 59.1 & 61.3 & 61.3 & 87.5 & 38.1 & 55.7 & 68.4 & 26.8 \\
VisionZip~\cite{yang2025visionzip} & MS &  58.8 & 61.3 & 61.2 & 85.3 & 35.8 & 54.6 & 68.5 & 26.3 \\
MustDrop~\cite{liu2024multi} & MS & 58.4 & 63.1 & 62.2 & 83.3 & 40.0 & 55.7 & 69.0 & 28.9 \\
VisionZip~\cite{yang2025visionzip} & PA+MS & 59.2 & 63.8 & 63.1 & 86.5 & 44.6 & 54.3 & 68.5 & 30.6 \\

\noalign{\hrule height 1pt}
\rowcolor{gray!20}
\multicolumn{10}{c}{\textit{Retain 128 Tokens in Average ($\downarrow$ 77.8\%)}} \\
VisionZip~\cite{yang2025visionzip} & PA & 57.9 & 62.5 & 62.8 & 84.4 & 44.3 & 54.5 & 68.6 & 29.9 \\
VisPruner~\cite{zhang2024beyond} & PS & 58.3 & 60.9 & 59.1 & 86.9 & 39.6 & 56.1 & 68.3 & 27.8 \\
DivPrune~\cite{alvar2025divprune} & PS & 59.4 & 62.2 & 61.3 & 87.4 & 41.9 & 56.1 & 68.8 & 28.6 \\
ToMe~\cite{bolya2022token} & MS & 58.4 & 60.6 & 60.0 & 87.2 & 36.0 & 56.7 & 68.3 & 24.6 \\
VisionZip~\cite{yang2025visionzip} & MS & 58.0 & 59.0 & 59.0 & 84.2 & 31.0 & 55.6 & 68.9 & 25.1 \\
MustDrop~\cite{liu2024multi} & MS & 56.8 & 60.7 & 57.7 & 81.5 & 34.3 & 56.6 & 67.8 & 27.2 \\
VisionZip~\cite{yang2025visionzip} & PA+MS & 57.6 & 62.2 & 63.3 & 84.7 & 44.2 & 54.8 & 68.7 & 29.9 \\

\noalign{\hrule height 1pt}
\rowcolor{gray!20}
\multicolumn{10}{c}{\textit{Retain 64 Tokens in Average ($\downarrow$ 88.9\%)}} \\
VisionZip~\cite{yang2025visionzip} & PA & 54.9 & 60.8 & 59.5 & 80.5 & 42.8 & 55.9 & 69.0 & 28.8 \\
VisPruner~\cite{zhang2024beyond} & PS & 56.3 & 58.2 & 57.9 & 85.9 & 35.6 & 56.7 & 67.3 & 24.4 \\
DivPrune~\cite{alvar2025divprune} & PS &57.8 & 59.1 & 57.8 & 86.3 & 39.4 & 57.6 & 68.2 & 28.3 \\
ToMe~\cite{bolya2022token} & MS & 56.6 & 58.8 & 57.5 & 85.2 & 31.1 & 57.3 & 67.7 & 21.0 \\
VisionZip~\cite{yang2025visionzip} & MS & 55.9 & 53.5 & 55.0 & 81.7 & 24.1 & 57.3 & 66.6 & 21.3 \\
MustDrop~\cite{liu2024multi} & MS & 53.3 & 47.1 & 49.2 & 77.9 & 16.5 & 56.4 & 64.1 & 17.8 \\
VisionZip~\cite{yang2025visionzip} & PA+MS & 55.2 & 60.0 & 60.4 & 80.9 & 42.2 & 55.5 & 68.8 & 28.4 \\

\noalign{\hrule height 1pt}
\end{tabular}
}

\caption{Detailed performance of different token reduction methods in Vision Tower (LLaVA-1.5-7B).}

\label{tab:supp_vision_llava_7b}
\end{table*}

\begin{table*}[t!]
\renewcommand{\arraystretch}{1.2}

\resizebox{\linewidth}{!}{
\begin{tabular}{lc|cccccccc}
\noalign{\hrule height 1pt}
\textbf{Method} & \textbf{Type} & \textbf{GQA} & \textbf{MMB} & \textbf{MME} & \textbf{POPE} & \textbf{TextVQA} & \textbf{VizWiz} & \textbf{ScienceQA} & \textbf{OCRBench} \\
\noalign{\hrule height 1pt}
\rowcolor{gray!20}

\multicolumn{10}{c}{\textit{Upper Bound, $\sim$2880 Tokens (100\%)}}\\
LLaVA-NeXT-7B~\cite{liu2024improved} & - &  64.3 & 67.1 & 66.0 & 87.6 & 64.7 & 60.7 & 70.1 & 52.1 \\
\noalign{\hrule height 1pt}
\rowcolor{gray!20}
\multicolumn{10}{c}{\textit{Retain $\sim$640 Tokens in Average ($\downarrow$ 77.8\%)}} \\
% ToMe~\cite{bolya2022token} & MS & 54.3 & 60.5 & -\\
VisionZip~\cite{yang2025visionzip} & PA & 61.9 & 65.4 & 63.0 & 87.6 & 61.9 & 59.9 & 67.8 & 48.8 \\
VisPruner~\cite{zhang2024beyond} & PS & 61.9 & 63.0 & 62.2 & 88.2 & 53.6 & 57.9 & 67.6 & 36.0 \\
DivPrune~\cite{alvar2025divprune} & PS & 61.4 & 64.7 & 65.0 & 87.2 & 54.4 & 58.8 & 67.5 & 37.6 \\
VisionZip~\cite{yang2025visionzip} & MS &  61.8 & 63.6 & 63.4 & 84.9 & 41.1 & 57.1 & 67.2 & 27.7 \\
MustDrop~\cite{liu2024multi} & MS & 62.7 & 65.6 & 62.9 & 85.5 & 49.8 & 58.2 & 69.4 & 32.9 \\

\noalign{\hrule height 1pt}
\rowcolor{gray!20}
\multicolumn{10}{c}{\textit{Retain $\sim$320 Tokens in Average ($\downarrow$ 88.9\%)}} \\
VisionZip~\cite{yang2025visionzip} & PA & 59.2 & 65.4 & 61.8 & 85.0 & 58.6 & 57.3 & 67.3 & 41.6 \\
VisPruner~\cite{zhang2024beyond} & PS & 59.2 & 63.0 & 58.9 & 86.9 & 47.7 & 55.4 & 68.2 & 28.1 \\
DivPrune~\cite{alvar2025divprune} & PS & 59.9 & 64.7 & 62.1 & 85.6 & 49.4 & 57.5 & 67.7 & 33.0 \\
VisionZip~\cite{yang2025visionzip} & MS & 58.9 & 63.6 & 58.1 & 82.9 & 34.3 & 55.6 & 67.7 & 23.0 \\
MustDrop~\cite{liu2024multi} & MS & 61.4 & 65.6 & 59.0 & 82.8 & 36.3 & 56.1 & 67.7 & 24.8 \\

\noalign{\hrule height 1pt}
\rowcolor{gray!20}
\multicolumn{10}{c}{\textit{Retain $\sim$160 Tokens in Average ($\downarrow$ 94.4\%)}} \\
VisionZip~\cite{yang2025visionzip} & PA & 54.9 & 65.4 & 56.5 & 81.0 & 54.3 & 55.8 & 67.7 & 33.1 \\
VisPruner~\cite{zhang2024beyond} & PS & 55.5 & 63.0 & 53.1 & 82.7 & 41.9 & 54.2 & 66.9 & 21.9 \\
DivPrune~\cite{alvar2025divprune} & PS &58.0 & 64.7 & 59.5 & 83.1 & 45.1 & 57.6 & 68.1 & 26.6 \\
VisionZip~\cite{yang2025visionzip} & MS & 55.8 & 63.6 & 50.0 & 77.8 & 25.0 & 54.3 & 66.4 & 15.4 \\

\noalign{\hrule height 1pt}
\end{tabular}
}

\caption{Detailed performance of different token reduction methods in Vision Tower (LLaVA-NeXT-7B).}

\label{tab:supp_vision_llava_next_7b}
\end{table*}

\begin{table*}[t!]
\renewcommand{\arraystretch}{1.2}

\resizebox{\linewidth}{!}{
\begin{tabular}{lc|cccccccc}
\noalign{\hrule height 1pt}
\textbf{Method} & \textbf{Type} & \textbf{GQA} & \textbf{MMB} & \textbf{MME} & \textbf{POPE} & \textbf{TextVQA} & \textbf{VizWiz} & \textbf{ScienceQA} & \textbf{OCRBench} \\
\noalign{\hrule height 1pt}
\rowcolor{gray!20}

\multicolumn{10}{c}{\textit{Upper Bound, 100\% Tokens}}\\
Qwen2.5-VL-7B & - & 60.4 & 82.9 & 83.3 & 87.4 & 82.9 & 70.8 & 87.4 & 83.7 \\
\noalign{\hrule height 1pt}
\rowcolor{gray!20}
\multicolumn{10}{c}{\textit{Retain 33.3\% Tokens in Average ($\downarrow$ 66.7\%)}} \\

VisionZip~\cite{yang2025visionzip} & PA & 58.9 & 82.8 & 81.9 & 86.7 & 78.7 & 69.4 & 86.7 & 73.4 \\
VisPruner~\cite{zhang2024beyond} & PS & 58.6 & 80.7 & 81.2 & 86.2 & 76.4 & 69.2 & 85.3 & 69.2 \\
DivPrune~\cite{alvar2025divprune} & PS & 59.6 & 81.4 & 80.0 & 86.6 & 78.9 & 69.6 & 84.9 & 72.9 \\
VisionZip~\cite{yang2025visionzip} & MS & 59.5 & 80.8 & 80.1 & 86.1 & 74.7 & 70.9 & 86.0 & 70.3 \\

\noalign{\hrule height 1pt}
\rowcolor{gray!20}
\multicolumn{10}{c}{\textit{Retain 22.2\% Tokens in Average ($\downarrow$ 77.8\%)}} \\
VisionZip~\cite{yang2025visionzip} & PA & 57.1 & 80.5 & 79.3 & 85.9 & 74.6 & 68.6 & 85.4 & 63.5 \\
VisPruner~\cite{zhang2024beyond} & PS & 57.3 & 79.1 & 77.6 & 84.6 & 70.2 & 67.9 & 84.4 & 58.1 \\
DivPrune~\cite{alvar2025divprune} & PS & 58.7 & 78.9 & 77.7 & 85.6 & 75.1 & 69.1 & 84.3 & 65.8 \\
VisionZip~\cite{yang2025visionzip} & MS & 58.4 & 79.6 & 79.5 & 85.4 & 66.0 & 69.5 & 83.8 & 59.8 \\

\noalign{\hrule height 1pt}
\rowcolor{gray!20}
\multicolumn{10}{c}{\textit{Retain 11.1\% Tokens in Average ($\downarrow$ 88.9\%)}} \\
VisionZip~\cite{yang2025visionzip} & PA & 52.6 & 76.9 & 70.0 & 82.9 & 62.9 & 67.1 & 82.9 & 42.8 \\
VisPruner~\cite{zhang2024beyond} & PS & 54.0 & 73.8 & 71.4 & 81.1 & 54.9 & 64.6 & 81.8 & 40.9 \\
DivPrune~\cite{alvar2025divprune} & PS & 56.2 & 75.7 & 72.4 & 82.4 & 65.5 & 66.4 & 82.1 & 50.1 \\
VisionZip~\cite{yang2025visionzip} & MS & 55.9 & 74.5 & 73.5 & 82.5 & 45.2 & 66.5 & 80.0 & 43.7 \\

\noalign{\hrule height 1pt}
\end{tabular}
}

\caption{Detailed performance of different token reduction methods in Vision Tower (Qwen2.5-VL-7B).}

\label{tab:supp_vision_qwen_7b}
\end{table*}

%% file: appendix/vision_tower_position.tex
\begin{table*}[t!]
\renewcommand{\arraystretch}{1.2}

\resizebox{\linewidth}{!}{
\begin{tabular}{lc|ccccccc|c}
\noalign{\hrule height 1pt}
\textbf{Method} & \textbf{Type} & \textbf{GQA} & \textbf{MMB} & \textbf{MME} & \textbf{POPE} & \textbf{TextVQA} & \textbf{VizWiz} & \textbf{ScienceQA} & \textbf{Prefill} \\
\noalign{\hrule height 1pt}
\rowcolor{gray!20}

\multicolumn{10}{c}{\textit{Upper Bound, 576 Tokens (100\%)}}\\
LLaVA-1.5-7B~\cite{liu2023visual} & - &  62.0 & 64.2 & 67.0 & 87.0 & 46.1 & 54.3 & 69.5 & 27.3 ms \\
\noalign{\hrule height 1pt}
\rowcolor{gray!20}
\multicolumn{10}{c}{\textit{Retain 192 Tokens in Average ($\downarrow$ 66.7\%)}} \\

ToMe\textsuperscript{*}~\cite{bolya2022token} & MS
 &  53.1 & 52.0 & 56.3 & 77.3 & 15.5 & 54.5 & 65.8 & 16.6 ms \\
ToMe\textsuperscript{\dag}~\cite{bolya2022token} & MS & 59.4 & 63.2 & 62.3 & 87.0 & 38.7 & 56.6 & 69.5 & 16.6 ms \\
MustDrop\textsuperscript{*}~\cite{liu2024multi} & MS & 56.9 & 60.2 & 56.9 & 83.9 & 30.8 & 54.9 & 67.7 & 16.6 ms \\
MustDrop\textsuperscript{\dag}~\cite{liu2024multi} & MS & 58.4 & 63.1 & 62.2 & 83.3 & 40.0 & 55.7 & 69.0 & 16.6 ms \\

\noalign{\hrule height 1pt}
\rowcolor{gray!20}
\multicolumn{10}{c}{\textit{Retain 128 Tokens in Average ($\downarrow$ 77.8\%)}} \\

ToMe\textsuperscript{*}~\cite{bolya2022token} & MS
 &  44.2 & 36.2 & 44.6 & 64.4 & 10.4 & 53.6 & 63.9 & 15.8 ms \\
ToMe\textsuperscript{\dag}~\cite{bolya2022token} & MS & 57.7 & 61.1 & 60.9 & 84.7 & 32.1 & 56.6 & 69.0 & 16.0 ms \\
MustDrop\textsuperscript{*}~\cite{liu2024multi} & MS & 52.6 & 52.8 & 53.2 & 78.0 & 22.7 & 54.5 & 65.6 & 16.0 ms \\
MustDrop\textsuperscript{\dag}~\cite{liu2024multi} & MS & 56.8 & 60.7 & 57.7 & 81.5 & 34.3 & 56.6 & 67.8 & 16.2 ms \\

\noalign{\hrule height 1pt}
\end{tabular}
}

\caption{Detailed results of token reduction at different layers in Vision Tower (LLaVA-1.5-7B). \textsuperscript{*} and \textsuperscript{\dag} denote merging in the first and last layer.}

\label{tab:supp_vision_tower_pos_llava_7b}
\end{table*}

%% file: appendix/llm.tex
\begin{table*}[t!]
\renewcommand{\arraystretch}{1.2}

\resizebox{\linewidth}{!}{
\begin{tabular}{lc|cccccccc}
\noalign{\hrule height 1pt}
\textbf{Method} & \textbf{Type} & \textbf{GQA} & \textbf{MMB} & \textbf{MME} & \textbf{POPE} & \textbf{TextVQA} & \textbf{VizWiz} & \textbf{ScienceQA} & \textbf{OCRBench} \\
\noalign{\hrule height 1pt}
\rowcolor{gray!20}

\multicolumn{10}{c}{\textit{Upper Bound, 576 Tokens (100\%)}}\\
LLaVA-1.5-7B~\cite{liu2023visual} & - &  62.0 & 64.2 & 67.0 & 87.0 & 46.1 & 54.3 & 69.5 & 31.3 \\
\noalign{\hrule height 1pt}
\rowcolor{gray!20}
\multicolumn{10}{c}{\textit{Retain 192 Tokens in Average ($\downarrow$ 66.7\%)}} \\

FastV~\cite{chen2024image} & PA & 58.4 & 63.5 & 64.5 & 83.7 & 43.5 & 54.4 & 69.2 & 28.5 \\
SparseVLM~\cite{zhang2024sparsevlm} & PA & 60.2 & 63.4 & 66.4 & 85.6 & 44.4 & 54.2 & 69.2 & 30.5 \\
DART~\cite{wen2025stop} & PS & 60.8 & 64.4 & 66.6 & 85.0 & 44.8 & 54.6 & 69.4 & 30.0 \\
HoliTom~\cite{shao2025holitom} & MS & 59.2 & 58.8 & 63.5 & 84.4 & 42.7 & 55.8 & 65.8 & 29.0 \\
SparseVLM~\cite{zhang2024sparsevlm} & PA+MS & 59.9 & 63.2 & 66.5 & 85.7 & 44.4 & 54.2 & 68.9 & 30.0 \\

\noalign{\hrule height 1pt}
\rowcolor{gray!20}
\multicolumn{10}{c}{\textit{Retain 128 Tokens in Average ($\downarrow$ 77.8\%)}} \\
FastV~\cite{chen2024image} & PA & 56.1 & 62.1 & 63.0 & 81.0 & 39.6 & 54.3 & 70.2 & 26.2 \\
SparseVLM~\cite{zhang2024sparsevlm} & PA & 58.6 & 63.1 & 65.1 & 84.1 & 42.2 & 54.6 & 69.4 & 28.3 \\
DART~\cite{wen2025stop} & PS & 59.5 & 63.8 & 65.4 & 83.9 & 43.5 & 54.7 & 69.7 & 29.8 \\
HoliTom~\cite{shao2025holitom} & MS & 57.8 & 57.0 & 62.0 & 82.8 & 39.2 & 55.4 & 66.4 & 26.6 \\
SparseVLM~\cite{zhang2024sparsevlm} & PA+MS & 58.6 & 63.2 & 65.1 & 84.7 & 42.5 & 54.3 & 69.6 & 27.9 \\

\noalign{\hrule height 1pt}
\rowcolor{gray!20}
\multicolumn{10}{c}{\textit{Retain 64 Tokens in Average ($\downarrow$ 88.9\%)}} \\

FastV~\cite{chen2024image} & PA & 52.2 & 58.3 & 57.8 & 75.1 & 32.1 & 53.4 & 68.6 & 18.8 \\
SparseVLM~\cite{zhang2024sparsevlm} & PA & 55.9 & 62.8 & 61.3 & 79.9 & 36.9 & 53.3 & 69.7 & 23.0 \\
DART~\cite{wen2025stop} & PS & 56.8 & 61.6 & 64.9 & 80.7 & 41.2 & 54.7 & 69.9 & 27.0 \\
HoliTom~\cite{shao2025holitom} & MS & 55.6 & 55.5 & 59.5 & 79.0 & 31.7 & 54.9 & 66.2 & 21.6 \\
SparseVLM~\cite{zhang2024sparsevlm} & PA+MS & 56.0 & 62.3 & 61.9 & 83.0 & 37.9 & 53.2 & 69.6 & 24.3 \\
\noalign{\hrule height 1pt}
\end{tabular}
}

\caption{Detailed performance of different token reduction methods in LLM (LLaVA-1.5-7B).}

\label{tab:supp_llm_llava_7b}
\end{table*}

\begin{table*}[t!]
\renewcommand{\arraystretch}{1.2}

\resizebox{\linewidth}{!}{
\begin{tabular}{lc|cccccccc}
\noalign{\hrule height 1pt}
\textbf{Method} & \textbf{Type} & \textbf{GQA} & \textbf{MMB} & \textbf{MME} & \textbf{POPE} & \textbf{TextVQA} & \textbf{VizWiz} & \textbf{ScienceQA} & \textbf{OCRBench} \\
\noalign{\hrule height 1pt}
\rowcolor{gray!20}

\multicolumn{10}{c}{\textit{Upper Bound, $\sim$2880 Tokens (100\%)}}\\
LLaVA-NeXT-7B~\cite{liu2024improved} & - &  64.3 & 67.1 & 66.0 & 87.6 & 64.7 & 60.7 & 70.1 & 52.1 \\
\noalign{\hrule height 1pt}
\rowcolor{gray!20}
\multicolumn{10}{c}{\textit{Retain $\sim$640 Tokens in Average ($\downarrow$ 77.8\%)}} \\

FastV~\cite{chen2024image} & PA & 61.0 & 64.8 & 63.0 & 86.1 & 58.6 & 59.0 & 68.2 & 38.9 \\
SparseVLM~\cite{zhang2024sparsevlm} & PA & 62.3 & 65.8 & 63.8 & 86.7 & 60.8 & 60.3 & 68.6 & 42.0 \\
DART~\cite{wen2025stop} & PS & 63.0 & 65.4 & 64.2 & 86.5 & 62.2 & 60.6 & 68.8 & 44.2 \\
HoliTom~\cite{shao2025holitom} & MS & 61.6 & 62.2 & 61.0 & 86.6 & 57.0 & 58.7 & 65.8 & 34.0 \\

\noalign{\hrule height 1pt}
\rowcolor{gray!20}
\multicolumn{10}{c}{\textit{Retain $\sim$320 Tokens in Average ($\downarrow$ 88.9\%)}} \\
FastV~\cite{chen2024image} & PA & 56.1 & 63.2 & 56.6 & 81.9 & 49.4 & 56.9 & 67.7 & 24.8 \\
SparseVLM~\cite{zhang2024sparsevlm} & PA & 59.7 & 64.8 & 62.7 & 85.0 & 53.8 & 59.1 & 68.9 & 30.4 \\
DART~\cite{wen2025stop} & PS & 60.7 & 63.9 & 62.5 & 84.2 & 58.2 & 59.4 & 67.9 & 35.4 \\
HoliTom~\cite{shao2025holitom} & MS & 58.3 & 59.4 & 58.5 & 83.4 & 48.9 & 56.1 & 65.9 & 24.8 \\

\noalign{\hrule height 1pt}
\rowcolor{gray!20}
\multicolumn{10}{c}{\textit{Retain $\sim$160 Tokens in Average ($\downarrow$ 94.4\%)}} \\
FastV~\cite{chen2024image} & PA & 51.0 & 56.1 & 50.9 & 74.2 & 38.5 & 55.1 & 65.9 & 15.4 \\
SparseVLM~\cite{zhang2024sparsevlm} & PA & 56.2 & 63.1 & 58.5 & 80.9 & 45.6 & 56.9 & 67.7 & 19.8 \\
DART~\cite{wen2025stop} & PS & 57.1 & 61.5 & 58.8 & 79.6 & 52.0 & 57.9 & 68.1 & 26.8 \\
HoliTom~\cite{shao2025holitom} & MS & 38.4 & 54.4 & 51.3 & 79.4 & 38.4 & 54.4 & 65.9 & 16.9 \\

\noalign{\hrule height 1pt}
\end{tabular}
}

\caption{Detailed performance of different token reduction methods in LLM (LLaVA-NeXT-7B).}

\label{tab:supp_llm_llava_next_7b}
\end{table*}

\begin{table*}[t!]
\renewcommand{\arraystretch}{1.2}

\resizebox{\linewidth}{!}{
\begin{tabular}{lc|cccccccc}
\noalign{\hrule height 1pt}
\textbf{Method} & \textbf{Type} & \textbf{GQA} & \textbf{MMB} & \textbf{MME} & \textbf{POPE} & \textbf{TextVQA} & \textbf{VizWiz} & \textbf{ScienceQA} & \textbf{OCRBench} \\
\noalign{\hrule height 1pt}
\rowcolor{gray!20}

\multicolumn{10}{c}{\textit{Upper Bound, 100\% Tokens}}\\
Qwen2.5-VL-7B & - & 60.4 & 82.9 & 83.3 & 87.4 & 82.9 & 70.8 & 87.4 & 83.7 \\
\noalign{\hrule height 1pt}
\rowcolor{gray!20}
\multicolumn{10}{c}{\textit{Retain 33.3\% Tokens in Average ($\downarrow$ 66.7\%)}} \\

FastV~\cite{chen2024image} & PA & 55.0 & 79.6 & 78.9 & 84.2 & 78.9 & 69.1 & 85.8 & 61.2 \\
SparseVLM~\cite{zhang2024sparsevlm} & PA & 57.0 & 82.0 & 80.9 & 86.9 & 80.9 & 69.8 & 86.0 & 63.8 \\
DART~\cite{wen2025stop} & PS & 58.4 & 80.9 & 80.6 & 84.9 & 77.5 & 68.9 & 85.0 & 65.7 \\
Holitom~\cite{shao2025holitom} & MS & 58.0 & 81.4 & 78.5 & 85.8 & 79.6 & 69.8 & 87.4 & 62.9 \\

\noalign{\hrule height 1pt}
\rowcolor{gray!20}
\multicolumn{10}{c}{\textit{Retain 22.2\% Tokens in Average ($\downarrow$ 77.8\%)}} \\
FastV~\cite{chen2024image} & PA & 51.4 & 75.3 & 74.4 & 81.6 & 75.1 & 67.3 & 82.2 & 49.1 \\
SparseVLM~\cite{zhang2024sparsevlm} & PA & 55.1 & 80.4 & 80.2 & 85.1 & 78.6 & 68.8 & 84.3 & 53.3 \\
DART~\cite{wen2025stop} & PS & 56.5 & 79.7 & 78.5 & 82.5 & 72.3 & 68.6 & 84.0 & 55.6 \\
Holitom~\cite{shao2025holitom} & MS & 56.7 & 78.2 & 76.5 & 84.8 & 76.3 & 68.0 & 84.7 & 55.3 \\

\noalign{\hrule height 1pt}
\rowcolor{gray!20}
\multicolumn{10}{c}{\textit{Retain 11.1\% Tokens in Average ($\downarrow$ 88.9\%)}} \\
FastV~\cite{chen2024image} & PA & 45.3 & 62.9 & 65.4 & 72.4 & 63.5 & 63.0 & 78.8 & 30.3 \\
SparseVLM~\cite{zhang2024sparsevlm} & PA & 51.5 & 74.1 & 76.2 & 81.6 & 71.2 & 66.8 & 82.4 & 31.4 \\
DART~\cite{wen2025stop} & PS & 52.2 & 73.7 & 73.4 & 77.5 & 61.1 & 65.3 & 81.5 & 43.6 \\
Holitom~\cite{shao2025holitom} & MS & 53.5 & 73.1 & 70.3 & 79.4 & 66.5 & 64.3 & 81.7 & 38.6 \\

\noalign{\hrule height 1pt}
\end{tabular}
}

\caption{Detailed performance of different token reduction methods in LLM (Qwen2.5-VL-7B).}

\label{tab:supp_llm_qwen_7b}
\end{table*}

\begin{table*}[t!]
\renewcommand{\arraystretch}{1.2}
\setlength{\tabcolsep}{5mm}
\resizebox{\linewidth}{!}{
\begin{tabular}{lc|cccc}
\toprule
\textbf{Model} & \textbf{Token Number} & \textbf{Prefill (ms)} & \textbf{Decode (ms)} & \textbf{Peak Memory (GB)} \\
\midrule
LLaVA-1.5-7B~\cite{liu2023visual} & 576 & 27.2 & 18.1 & 13.9 \\
\midrule
ToMe~\cite{bolya2022token} & 128 & 17.7 & 17.6 & 13.4 \\
VisionZip~\cite{yang2025visionzip} & 128 & 17.5 & 17.4 & 13.5 \\
DivPrune~\cite{alvar2025divprune} & 128 & 17.3 & 17.6 & 13.4 \\
FastV~\cite{chen2024image} & 128 & 19.1 & 17.7 & 13.5 \\
DART~\cite{wen2025stop} & 128 & 22.9 & 17.7 & 13.5 \\
\midrule
LLaVA-1.5-13B~\cite{liu2023visual} & 576 & 44.6 & 17.9 & 26.1 \\
\midrule
ToMe~\cite{bolya2022token} & 128 & 22.6 & 17.6 & 25.4 \\
VisionZip~\cite{yang2025visionzip} & 128 & 23.0 & 17.6 & 25.4 \\
DivPrune~\cite{alvar2025divprune} & 128 & 21.5 & 17.6 & 25.3 \\
FastV~\cite{chen2024image} & 128 & 25.8 & 17.6 & 25.5 \\
DART~\cite{wen2025stop} & 128 & 27.4 & 17.5 & 25.4 \\
\bottomrule
\end{tabular}
}

\caption{The actual latency and memory usage profiles of different algorithms.}

\label{tab:token_performance}
\end{table*}